\documentclass[10pt,twocolumn,letterpaper]{article}

\usepackage[pagenumbers]{cvpr} 
\usepackage{graphicx}
\usepackage{multirow}
\usepackage{pifont}
\usepackage{algorithm}
\usepackage{algpseudocode}
\usepackage[normalem]{ulem}
\usepackage{amsmath}
\usepackage{amsthm}
\newtheorem{lemma}{Lemma}
\usepackage{wrapfig}
\usepackage{colortbl}
\usepackage[utf8x]{inputenc}
\usepackage{cuted}

\usepackage[dvipsnames]{xcolor}

\definecolor{cvprblue}{rgb}{0.21,0.49,0.74}
\usepackage[pagebackref,breaklinks,colorlinks,citecolor=cvprblue]{hyperref}

\newcommand{\paravspace}{\vspace{-10pt}}

\def\paperID{2006} 
\def\confName{CVPR}
\def\confYear{2025}

\title{\textit{MoGe}: Unlocking Accurate Monocular Geometry Estimation for Open-Domain Images with Optimal Training Supervision}

\makeatletter
\renewcommand*{\@fnsymbol}[1]{\ensuremath{\ifcase#1\or \star\or \dagger\or \ddagger\or
		\mathsection\or \mathparagraph\or \|\or **\or \dagger\dagger
		\or \ddagger\ddagger \else\@ctrerr\fi}}
\makeatother

\author{\!\!Ruicheng Wang$^{1}$\!\ \!\thanks{Work done during internship at Microsoft Research}\,\, \  Sicheng Xu$^{2}$  \ Cassie Dai$^{3\star}$ \ Jianfeng Xiang$^{4\star}$ \ Yu Deng$^{2}$ \ Xin Tong$^{2}$ \ Jiaolong Yang$^{2}$\thanks{Corresponding author} \\
	$^1${USTC} \quad $^2${Microsoft Research} \quad $^3${Harvard} \quad $^4${Tsinghua University} \\
	{\tt\small \{t-ruiwang,sichengxu,t-jxiang,dengyu,jiaoyan\}@microsoft.com  \{cccassied,xtong.gfx\}@gmail.com}
 \vspace{-95pt}
}

\begin{document}

\maketitle

\begin{strip}
	\centering
	\includegraphics[width=0.92\textwidth]{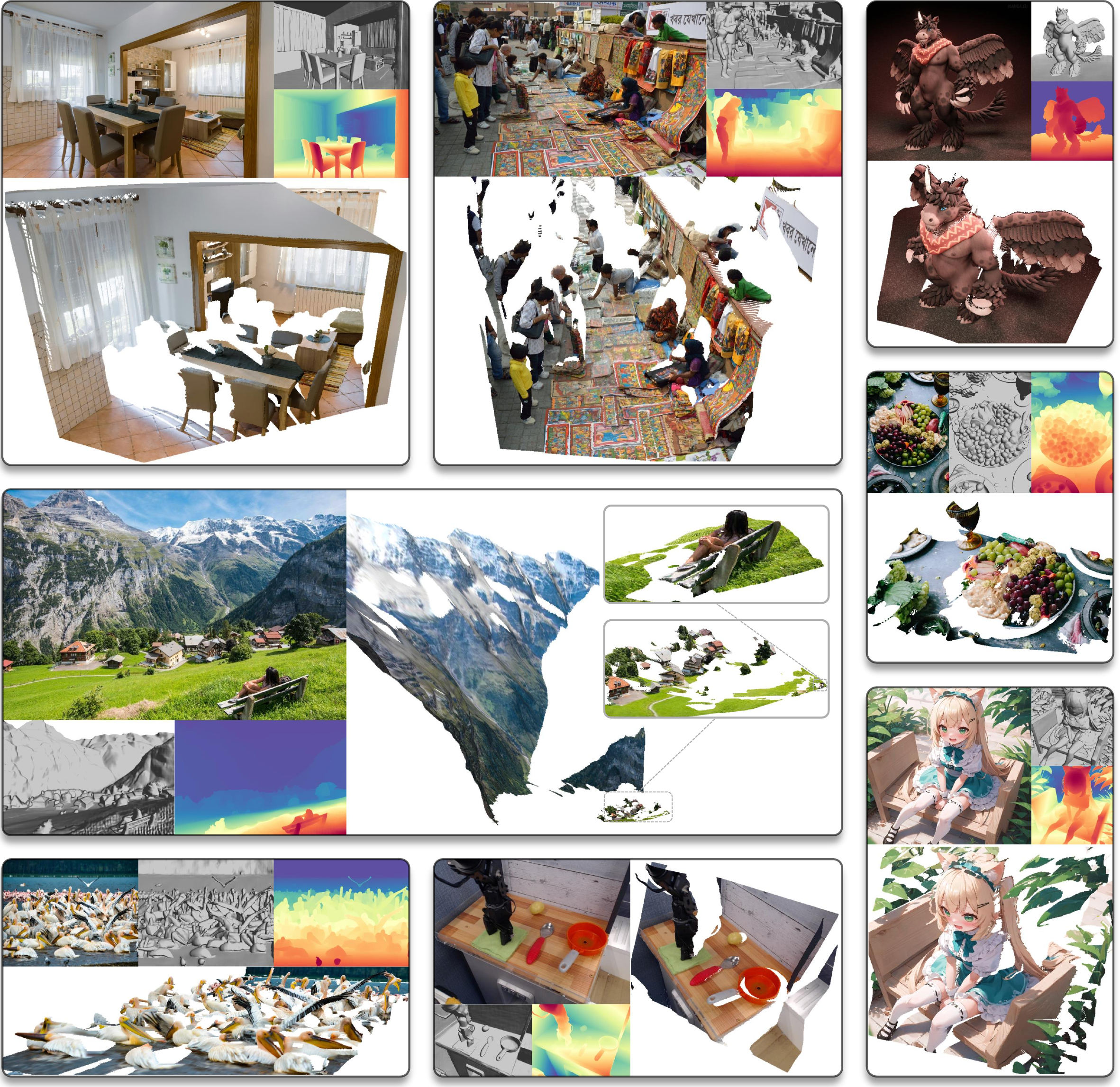}
	\captionof{figure}{
        Given any image, our method reconstructs an affine-invariant 3D point map of the scene and can also produce a depth map and the camera focal length. The model yields high-quality shapes and generalizes well to open-domain images. (\textit{Best viewed with zoom})
    }
    \label{fig:teaser}
	\vspace{-5pt}
\end{strip}

\setlength{\abovedisplayskip}{5.5pt} 
\setlength{\belowdisplayskip}{5.5pt} 

\begin{abstract}
We present MoGe, a powerful model for recovering 3D geometry from monocular open-domain images. 
Given a single image, our model directly predicts a 3D point map of the captured scene with an affine-invariant representation, which is agnostic to true global scale and shift. This new representation precludes ambiguous supervision in training and facilitates effective geometry learning. Furthermore, we propose a set of novel global and local geometry supervision techniques that empower the model to learn high-quality geometry. These include a robust, optimal, and efficient point cloud alignment solver for accurate global shape learning, and a multi-scale local geometry loss promoting precise local geometry supervision. We train our model on a large, mixed dataset and demonstrate its strong generalizability and high accuracy. In our comprehensive evaluation on diverse unseen datasets, our model significantly outperforms state-of-the-art methods across all tasks, including monocular estimation of 3D point map, depth map, and camera field of view. Code and models can be found on our \href{https://wangrc.site/MoGePage/}{project page}.

\end{abstract}

\section{Introduction}

Estimating the 3D geometry of general scenes is a fundamental task in computer vision. While 3D reconstruction from a set of images has been extensively studied with Structure-from-Motion (SfM)~\cite{agarwal2011building, schonberger2016structure} and Multi-View Stereo (MVS)~\cite{schonberger2016pixelwise,yang2020cost} techniques, recovering the 3D geometry from a single image in an arbitrary domain remains a significant challenge due to its highly ill-posed nature. 

For monocular geometry estimation (MGE), a common approach involves  first estimating a depth map, subject to an unknown scale (and shift), and then combining it with camera intrinsics to recover the 3D shape via unprojection~\cite{wei2021leres, piccinelli2024unidepth}. For the former, significant advancements in monocular depth estimation (MDE) have been achieved in recent years by training on large-scale data~\cite{yin2023metric3d, hu2024metric3dv2, yang2024depthanything, yang2024depthanythingv2, ranftl2020midas} or leveraging powerful pretrained models~\cite{ke2024marigold,fu2024geowizard,gui2024depthfm}. However, inferring camera intrinsics (\eg, focal length) from single images remains challenging due to the high degree of ambiguity when strong geometric cues are absent. Inaccurate camera parameters can lead to significant geometry distortions even when used with ground truth depth maps. Recently, DUSt3R~\cite{wang2024dust3r} proposed predicting a 3D point map from an image by mapping each pixel to a free 3D point, thereby bypassing camera parameter estimation during geometry recovery. Although the model can estimate monocular geometry, it is primarily designed for multi-view scenarios and is trained on image pair inputs. 

In this paper, we introduce a new \textit{direct} 3D geometry estimation method designed for single images in the open domain. Our model architecture is simple and straightforward: it directly predicts point maps from images, which can further derive depth map and camera focal length or FOV if needed. Different from DUSt3R~\cite{wang2024dust3r} which uses scale-invariant point maps, our method predicts \textit{affine-invariant} point maps, where the 3D points are subject to an unknown global scale as well as a 3D shift.  
This alteration is important as it eliminates the focal-distance ambiguity which is detrimental to the network training -- we provide intuitive explanations with illustration as well as empirical evidence to support this advancement.

\begin{figure}[t!]
	\centering
	\includegraphics[width=\linewidth]{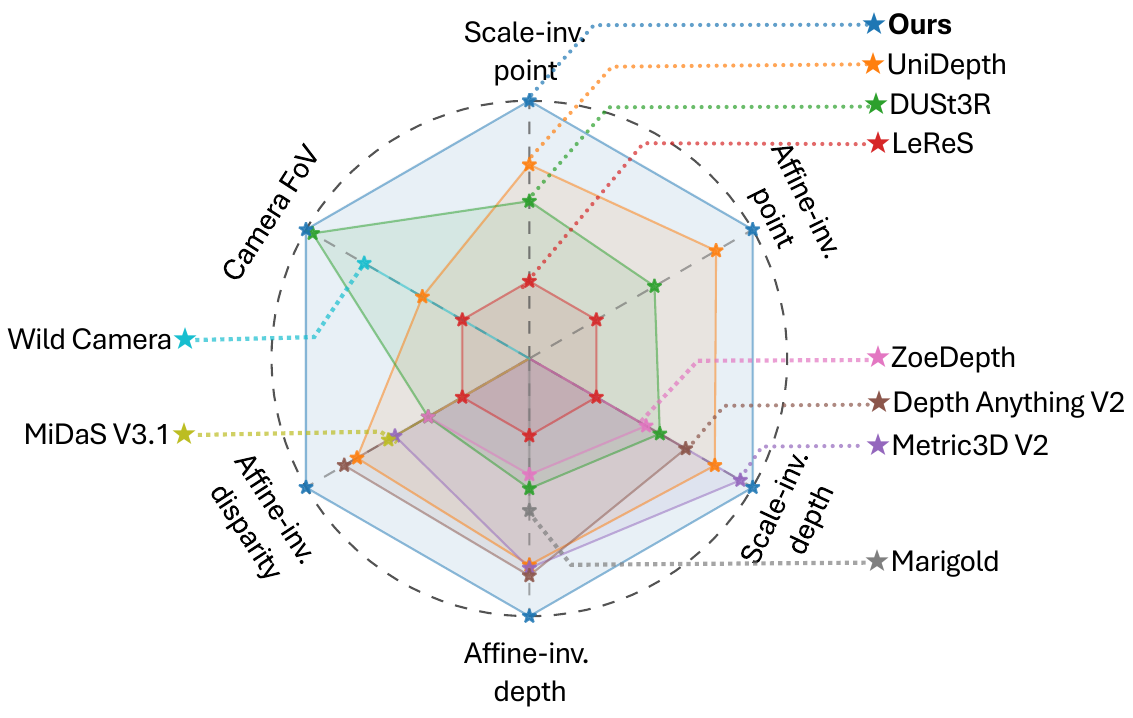}
    \vspace{-15pt}
	\caption{
		Accuracy ranking of existing methods and ours in our extensive evaluation on monocular 3D point map (scale-invariant and affine-invariant), depth map (scale-invariant depth, affine-invariant depth, and affine-invariant disparity), and camera field of view. Outer methods rank higher. See our experiments for details.
		}
	\label{fig:performance}
    \vspace{-15pt}
\end{figure}

More importantly, our findings indicate that \emph{the design of training supervisions is crucial} to achieve robust and accurate point map prediction from single images. Similar to previous MDE approaches, a global scaling factor and translation are required to align predictions with ground truth during training. However, existing methods for this global alignment calculation are either sensitive to outliers or solved with coarse approximations, leading to suboptimal supervision. We propose a \textit{Robust, Optimal, and Efficient (ROE)} global alignment solver to resolve scale and shift for the affine-invariant point map loss, which substantially improves the training effectiveness and final accuracy. On the other hand, the effective learning of local, region-specific geometry has been largely neglected previously. In monocular geometry estimation, the relative distances between different objects can be ambiguous, which hinders precise local geometry learning when only a global alignment is applied. In light of this, we propose a \textit{multi-scale local geometry loss}, which penalizes local discrepancies in 3D point clouds under independent affine alignments. This design significantly boosted the accuracy of our local geometry prediction.

We train our model on large-scale data sourced from various existing datasets. Our model demonstrates strong generalizability and accuracy in monocular geometry estimation for open-domain images (examples in Fig.~\ref{fig:teaser}). Zero-shot evaluation on eight unseen datasets shows a significant improvement over previous MGE methods that output point clouds (over \textbf{35\%} error reduction compared to the previous best). Moreover, our method significantly outperforms prior approaches that focus on the sub-tasks of MDE and camera FOV (\textbf{20\%$\sim$30\%} error reduction for the former and over \textbf{20\%} for the latter). Figure~\ref{fig:performance} shows that our method ranks top across all tasks and metrics in the evaluation.

Our contributions are summarized as follows: 1) We propose a new direct MGE method for open-domain images using affine-invariant point maps. 2) We establish a set of novel and effective global and local supervisions for robust and precise geometry recovery. 3) We demonstrate the substantial enhancement in performance of our model over existing methods on MGE, MDE, and camera FOV estimation across diverse datasets. 
We hope that our method serves as a versatile foundational model for monocular geometry tasks, facilitating applications such as 3D-aware image editing, depth-to-image synthesis, novel view synthesis, and 3D scene understanding, as well as providing initial geometrical priors for video or multiview-based 3D reconstruction.

\section{Related Work}

\paragraph{Monocular depth estimation.}
As a crucial precursor to monocular geometry estimation, monocular depth estimation is a long-standing task and has been extensively studied in the past. A number of   methods~\cite{eigen2014depth,bhat2021adabins,yin2023metric3d,hu2024metric3dv2,piccinelli2024unidepth,bhat2023zoedepth,li2024binsformer} have been developed to estimate depth with metric scale, which often rely heavily on data from specific sensors such as RGBD cameras, LiDAR, and calibrated stereo cameras. This dependence restricts their applicability to specific domains, such as indoor and driving scenes. In contrast, relative depth estimation has garnered significant attention due to its ability to utilize a much wider variety of data, thereby exhibiting enhanced generalizability. A prevalent approach is to predict relative depth in an affine-invariant manner, either through direct regression~\cite{chen2016diw,MegaDepthLi18,godard2019monodepth,chen2020oasis,yin2020diversedepth,ranftl2020midas,yang2024depthanything, yang2024depthanythingv2} or generative modeling~\cite{ke2024marigold,fu2024geowizard,wan2023metaprompt,gui2024depthfm}. 
Despite the recent advancements in monocular depth estimation, recovering 3D shape from depth information always necessitates known camera intrinsics.

\paravspace
\paragraph{Monocular point map estimation.}
Compared to depth maps, 3D point map offers a more straightforward representation of 3D geometry.
Monocular point map estimation aims to recover free 3D points for each pixel in a single image. Several approaches address this by predicting camera parameters alongside depth.
For example, LeReS~\cite{wei2021leres,yin2022accurate3dscene} introduced a two-stage pipeline with affine-invariant depth prediction followed by a point cloud module to recover shift and camera focal length.
UniDepth~\cite{piccinelli2024unidepth} proposed a self-promptable camera module that predicts dense camera representation to condition the subsequent depth estimation module. 
DUSt3R~\cite{wang2024dust3r} employs an end-to-end model to directly map two-view images to camera space point maps, adaptable to monocular scenarios by using two identical input images. However, its scale-invariant point map can suffer from the focal-distance ambiguity. 
In contrast, our approach is designed to handle monocular input and we employ affine-invariant point map with meticulously designed training supervisions for more effective geometry learning.

\paravspace
\paragraph{Camera intrinsics estimation.}
Estimating camera intrinsics is a fundamental task in 3D vision.
While traditional multiview camera calibration using special patterns is well studied~\cite{zhang2000flexible}, estimating camera intrinsics from single images is ill-posed and remains challenging. For the latter, earlier works utilized known 3D shapes~\cite{wilczkowiak2001camera} or vanishing points~\cite{deutscher2002manhattan}. 
More recently, learning-based approaches have been explored \cite{workman2015deepfocal,bogdan2018deepcalib,lee2021ctrlc,jin2023perspectivefields,zhu2024wildcamera} to handle in-the-wild images, but the results are far from satisfactory. In this work, we demonstrate that our camera parameter estimation derived from point maps can achieve state-of-the-art accuracy and offer remarkable generalizability.

\paravspace
\paragraph{Large-scale data training for monocular geometry.}
Recently, there has been an emerging trend~\cite{ranftl2020midas,eftekhar2021omnidata,bhat2023zoedepth,piccinelli2024unidepth,yin2023metric3d,hu2024metric3dv2,wang2024dust3r,yang2024depthanything,yang2024depthanythingv2} 
in monocular geometry estimation and depth estimation that leverages large-scale datasets coupled with advanced network backbones~\cite{dosovitskiy2020vit, ranftl2021dpt, oquab2023dinov2} to enhance performance. MiDaS~\cite{ranftl2020midas} marks an important milestone by mixing large datasets from various domains for training. 
Depth Anything~\cite{yang2024depthanything} leverages both labeled and extensive unlabeled data to improve generalization. Its successive work~\cite{yang2024depthanythingv2} further highlights the importance of high-quality synthetic data for capturing rich shape details in depth prediction. These advancements underscore the vital importance of large-scale training data in improving the performance of monocular geometry estimation.

\begin{figure*}[ht!]
	\centering
	\includegraphics[width=\linewidth]{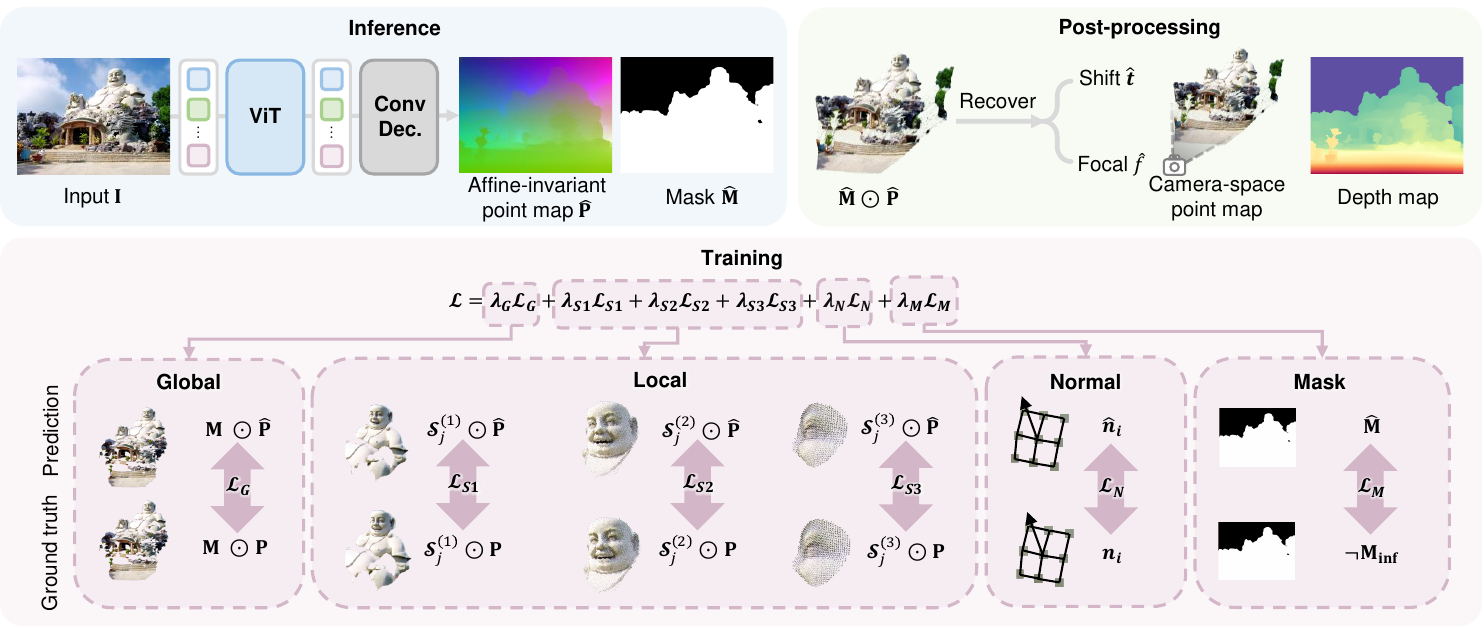}
	\vspace{-18pt}
	\caption{Method overview. Our model consists of a ViT encoder and a convolutional decoder. It predicts an affine-invariant point map as well as a mask that excludes regions with undefined geometry (\eg,  infinity). Depth, camera shift, and focal length can be further derived from the model output. For training, we design robust and effective supervisions focusing on both the global and local geometry.}
	\label{fig:overview}
\end{figure*}

\vspace{-5pt}
\begin{figure*}[htbp]  
	\centering  
	\begin{minipage}[t]{0.775\textwidth}  
		\centering
		\includegraphics[width=\linewidth]{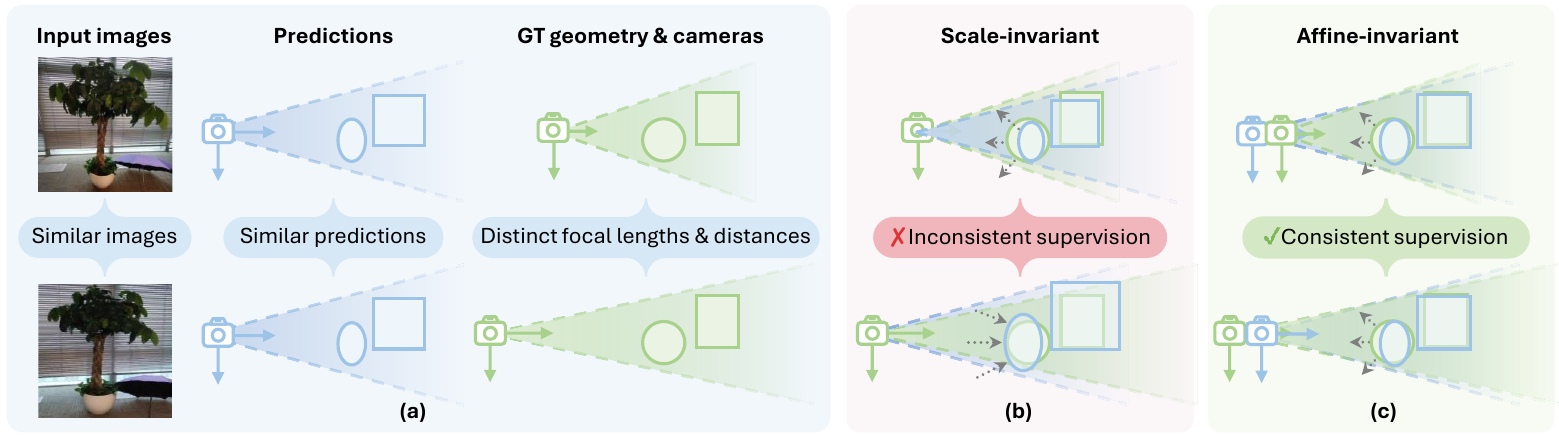}
		\vspace{-18pt}
		\caption{The focal-distance ambiguity and effects of different 3D point representations. (a) For similar images captured with varying camera focal length and distance to the objects, perceiving their true camera setup is challenging and models often produce similar geometries. (b) Inconsistent supervision signals occur with only scale alignment. (c) Consistent geometry supervision with an additional translation alignment.}
		\label{fig:affine}
	\end{minipage}  
	\hfill  
	\begin{minipage}[t]{0.2015\textwidth}  
		\vspace{-107.5pt}
		\includegraphics[width=\textwidth]{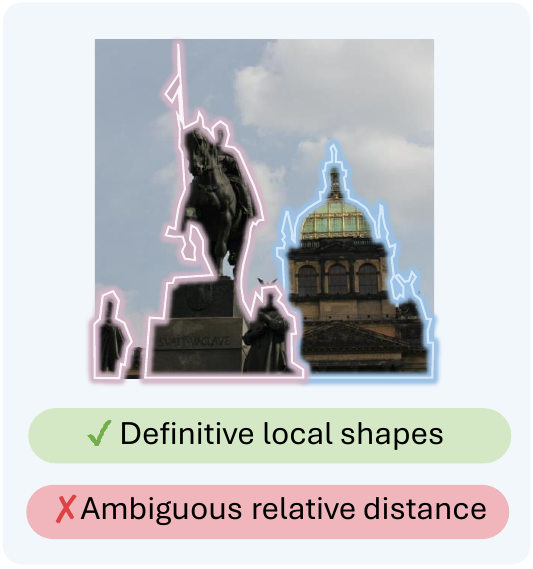}
		\vspace{-17.5pt}
		\caption{The relative distance ambiguity, where distances between objects can be hard to predict.}
		\label{fig:relative_distance_ambiguity}
	\end{minipage}   
    \vspace{-15pt}
\end{figure*}

\section{Approach}
Our method takes a single image as input and directly predicts the geometry of the scene represented by 3D points. An overview of the method is shown in Fig.~\ref{fig:overview}.

\subsection{Affine-invariant point map}
For an image $\mathbf I\in\mathbb R^{H\times W\times 3}$, our model $F_\theta$ infers the 3D coordinates of the image pixels, represented by a point map $\mathbf P\in\mathbb R^{H\times W \times 3}$, \ie,:
\begin{equation}
    F_\theta:\mathbf I\mapsto \mathbf P.
\end{equation}
The $X$ and $Y$ axes of the coordinate frame of $\mathbf P$ are aligned with the $u$ and $v$ axes in the image space respectively.

Monocular geometry estimation often suffers from focal-distance ambiguity, as shown in Fig.~\ref{fig:affine}~(a). To deal with this issue, we propose to predict \textit{affine-invariant} point map $\mathbf P$, \ie, $\mathbf P$ is agnostic to the global scale $s\in \mathbb R$ and offset $\mathbf t\in \mathbb R^3$ and thus $\mathbf P \cong s\mathbf P+\mathbf t, \forall s, \forall \mathbf t$. 
Compared to the scale-invariant supervision~\cite{wang2024dust3r} shown in Fig.~\ref{fig:affine}~(b), the discrepancies in focal lengths and camera positions between the predictions and ground truth are amended by a translation, as shown in Fig.~\ref{fig:affine}~(c). This ensures consistent geometry supervision when focal length is ambiguous, enabling more effective geometry learning.

Following accepted practice, we assume that the camera principal point coincides with image center and pixels are square. In this case, $\mathbf{t}$ can be simplified with a $Z$-axis shift $t_z$ (\ie, $t_x = t_y=0$).

\paravspace
\paragraph{Recovering camera focal and shift.} 
The affine-invariant point map representation can be used to recover both camera shift and focal length. Given the predicted 3D points  $(x_i, y_i, z_i)$ and their corresponding 2D pixels $(u_i, v_i)$, we solve for camera focal length prediction $f$ and $Z$-axis shift $t_z'$ by minimizing the projection error: 
\begin{equation}
    \min_{f, t_z'} \sum_{i=1}^N \left({f x_i\over z_i +  t_z'} - u_i \right)^2 + \left({f y_i\over z_i + t_z'} - v_i \right)^2,
    \label{eq:reover_shift_focal}
\end{equation}
where  $t_z'=t_z/s$. 
Eq.~\ref{eq:reover_shift_focal} can be efficiently solved with a fast iterative optimization algorithm, which typically converges within 10 iterations, taking approximately $3$ms. More details about the solution can be found in the \emph{suppl. materials}. 

With $t_z'$ recovered, the scale-invariant depth map and point map in the camera space can be obtained by adding it to the $z$ coordinates.

\subsection{Training Objectives}
\label{sec:training}
\paragraph{Global point map supervision.} 
Let $\hat{\mathbf{p}}_i$ denote the predicted 3D point for the $i$-th pixel, and $\mathbf p_i$ the corresponding ground truth. The global point map loss is defined as:
\begin{equation}
        \mathcal L_G = \sum_{i\in {\mathcal M}} {\frac{1}{{z}_i}\| s\hat{\mathbf{p}}_i + \mathbf t-{\mathbf p}_i\|_1},
    \label{eq:global_loss}
\end{equation}
where $\|\!\cdot\!\|_1$ denotes the $L_1$ norm, $s$ and $\mathbf t$ are the alignment parameters that transform the predicted affine-invariant point map to the ground-truth camera space, and ${\mathcal M}$ is the mask for regions with labels. The weighting term $\frac{1}{{z}_i}$, where $z_i$ is the $z$-coordinate of ${\mathbf{p}}_i$, is applied to balance the supervision signal across extreme depths variations. 

To apply the global loss $\mathcal{L}_G$ for training, we need to first determine $s$ and $\mathbf t$. Previous affine-invariant depth estimation methods often use rough alignment approximations, like using the median depth of two sets as an anchor to compute shift, followed by scaling \cite{ranftl2020midas,yang2024depthanything,yang2024depthanythingv2}, or  normalizing depth range \cite{ke2024marigold,fu2024geowizard}. These simple strategies yield suboptimal alignment and could lead to unsatisfactory supervision. 

In this work, we propose a solver to the {\emph{optimal}} alignment parameters. Specifically, we determine $s^*$ and $\mathbf t^*$ via
\begin{equation}
\begin{split}
    (s^*,\mathbf t^*)=\mathop{\arg\!\min}_{s,\mathbf t}\sum_{i\in {\mathcal M}} \frac{1}{{z}_i}\left\|s\hat{\mathbf p}_i + \mathbf t-{\mathbf p}_i\right\|_1,
\end{split}
\label{eq:global_align}
\end{equation}
where $t_x$ and $t_y$ are $0$. 
One approach to solve this equation is to frame it as an absolute residuals optimization problem and use linear programming methods such as simplex algorithms and the interior-point method \cite{boyd2004convex}. However, these methods are known to have high computational complexity, often exceeding $O(N^3)$, making them inefficient for network training with thousands of points, where computation can take seconds.
Instead, we develop an efficient parallelized searching algorithm. With the fact that the optimum must occur when $s\hat z_k + t_z = z_k$ for some index $k$, we replace $t_z$ with $s$ and break the problem into a series of parallel one-dimensional subproblems. This  reduces the complexity to $O(N^2 \log N)$ and enables efficient GPU-based training. A high-level description of this solution is presented in Algorithm~\ref{alg:solver} with more details in the \emph{suppl. materials}.

In practice, we found that Eq.~\ref{eq:global_align} is still sensitive to outliers occasionally even with the $L_1$ error norm. If the model mistakenly predicts a close foreground edge pixel as a distant background point, the objective would be dominated by this incorrect prediction. 
To further improve the robustness, we apply truncation $\min(\cdot, \tau)$ to the absolute residuals. While this truncation introduces non-convexity, our solver remains applicable with minor adjustments.

In summary, our alignment solver is made \emph{Robust}, \emph{Optimal}, and \emph{Efficient}, and hence we call it the \emph{ROE} solver. 

\begin{figure}[t]
\vspace{-0.6em}
\begin{algorithm}[H]
\caption{Overview of ROE alignment}
\label{alg:solver}
\begin{algorithmic}    
	\small
	\Function{SolveSubproblem}{\null}
	\State{Enumerate breakpoints of the piecewise linear function.}
	\State{Find extrema among the breakpoints by their derivatives.}
	\State \Return{$s^{(k)}$ with smallest objective value $l^{(k)}$ at extrema.}
	\EndFunction
	\vspace{-6pt}
	\Statex
	\For{index $k = 1$ to $N$} \Comment{\emph{parallel computation}} 
	\State Formulate subproblem by substituting $t_z$ with $z_k\!-s^{(k)}\hat{z}_k$.
	\State{Solve scale $s^{(k)}$ and $l^{(k)}$ via \Call{SolveSubproblem}{\null}.}
	\State Obtain translation $t_z^{(k)}$ as $z_k\!-\!s^{(k)}\hat{z}_k$.
	\EndFor
	\State Select optimal $s^*$ and $t_z^*$ with smallest function value $l^{(k)}$.
 \end{algorithmic}
\end{algorithm}
\vspace{-25pt}
\end{figure}

\paravspace
\paragraph{Multi-scale local geometry loss.}
\label{local_loss}

In monocular geometry estimation, the relative distance between different objects can be ambiguous and difficult to predict, as illustrated in Fig.~\ref{fig:relative_distance_ambiguity}. This hinders precise local geometry learning when a global alignment computed with all objects is applied. To enhance the supervision for local geometry, we propose a loss function which measures the accuracy of local sphere regions with \emph{independent} alignment.

Specifically, given a ground-truth 3D point ${\mathbf p}_j$ as the anchor, we first select the point set within the spherical region centered at ${\mathbf p}_j$, defined as:
\begin{equation}
	{\mathcal S}_j = \{ i \mid \|{\mathbf p}_i - {\mathbf p}_j\| \leq r_j,i\in {\mathcal M} \},
\end{equation}
where  $r_j$ is the radius. We set $r_j = \alpha\cdot z_{j}\cdot\frac{\sqrt{W^2+H^2}}{2 \cdot f}$ where $z_j$ is the $z$-coordinate of $p_j$, $f$ is the ground truth focal length, and $W$ and $H$ are image width and height. This way, the hyper-parameter $\alpha\in(0,1)$ approximates the proportion of the projected sphere's diameter to the image diagonal. 

Then, we apply the aforementioned ROE alignment solver to align the point maps with optimal $(s_j^*,\mathbf t_j^*)$ and compute error. For each sphere scale parameter $\alpha$, we sample a set of anchor points $\mathcal H_\alpha$ and compute the loss as 
\begin{equation}
    \mathcal L_{S(\alpha)} =\! \sum_{j\in \mathcal H_\alpha} l_{{\mathcal S}_j} \!=\! \sum_{j\in \mathcal H_\alpha}\! \sum_{i\in { S}_j} \frac{1}{z_i}\left\| s_j^* \hat{\mathbf p}_i+\mathbf t_j^*- {\mathbf p}_i\right\|_1.
\end{equation}
In practice, we use three  $\alpha$ scales from coarse to fine: $1/4$, $1/16$, and $1/64$, denoted as $\mathcal L_{S_1}$, $\mathcal L_{S_2}$ and $\mathcal L_{S_3}$, respectively.

\paravspace
\paragraph{Normal loss.}
For better surface quality, we additionally supervise the normal computed from the predicted point map with the ground truth:
\begin{equation}
    \mathcal L_N = \sum_{i\in {\mathcal M}}\angle (\hat{\mathbf n}_i, \mathbf n_i),
\end{equation}
where the normal $\hat{\mathbf n}_i$ of pixel $i$ is obtained from the cross product of its adjacent edges on the image grid, $\mathbf n_i$ is the ground truth, and $\angle (\cdot,\cdot)$ measures the angle difference between the two vectors.

\paravspace
\paragraph{Mask loss.}
The infinity region (\eg, sky)  in outdoor scenes and plain backgrounds in object-only images have undefined geometry. 
We use a single-channel output head for predicting the mask $\hat{\mathbf M}\in \mathbb R^{H\times W}$ of valid points:
\begin{equation}
    \mathcal L_M = \|\hat{\mathbf M} - (1-{\mathbf M}_\text{inf})\|_2^2,
\end{equation}
where ${\mathbf M}_\text{inf}$ is the infinity mask label. For synthetic data, the ground truth masks are readily available, while for real outdoor scenes, we use SegFormer~\cite{xie2021segformer} to obtain sky masks. At inference time, the predicted mask is binarized with a threshold of $0.5$. The recovery of focal and shift involves masked points, excluding the regions of invalid geometry.

\subsection{Training data}

We collected 21 publicly available datasets from various domains~\cite{geyer2020a2d2, Argoverse2, dehghan2021arkitscenes, diml, yao2020blendedmvs, MegaDepthLi18, zamir2018taskonomy, sun2020waymo, Wang2019gtasfm, roberts2021hypersim, wang2019IRS, niklaus2019kenburns, li2023matrixcity, Fonder2019MidAir, huang2018mvsynth, Mehl2023Spring, Structured3D, Ros2016synthia, tartanair2020iros, gómez2023urbansyn, objaverse}, totaling around 9 million frames for training our model. 
Since the quality of data labels are varied, we apply tailored loss function combinations for different data sources. Higher-quality and less noisy data are assigned with finer-level loss functions. 
To address imperfections and potential outliers in real labels, we exclude the top $5\%$ of per-pixel loss values for real data. Following DINOv2~\cite{oquab2023dinov2}, we balance the weights of collected datasets via image retrieval from a curated image set. 
Details of training datasets can be found in the \emph{suppl. materials}.

\section{Experiments}

\begin{table*}[ht!]
    \footnotesize
    \centering
    \setlength{\tabcolsep}{4pt}
    \begin{tabular}{l|cc|cc|cc|cc|cc|cc|cc|cc|ccc}
        \specialrule{.12em}{0em}{0em}
        \multirow{2}{*}{\textbf{Method}} & \multicolumn{2}{c|}{NYUv2} & \multicolumn{2}{c|}{KITTI} & \multicolumn{2}{c|}{ETH3D} & \multicolumn{2}{c|}{iBims-1} & \multicolumn{2}{c|}{GSO} & \multicolumn{2}{c|}{Sintel} & \multicolumn{2}{c|}{DDAD} & \multicolumn{2}{c|}{DIODE} & \multicolumn{3}{c}{Average}\\
        & \scriptsize Rel\textsuperscript{p}\tiny$\downarrow$ & \scriptsize $\delta_1^\text{p}$\tiny$\uparrow$ & \scriptsize Rel\textsuperscript{p}\tiny$\downarrow$ & \scriptsize $\delta_1^\text{p}$\tiny$\uparrow$ & \scriptsize Rel\textsuperscript{p}\tiny$\downarrow$ & \scriptsize $\delta_1^\text{p}$\tiny$\uparrow$ & \scriptsize Rel\textsuperscript{p}\tiny$\downarrow$ & \scriptsize $\delta_1^\text{p}$\tiny$\uparrow$ & \scriptsize Rel\textsuperscript{p}\tiny$\downarrow$ & \scriptsize $\delta_1^\text{p}$\tiny$\uparrow$ & \scriptsize Rel\textsuperscript{p}\tiny$\downarrow$ & \scriptsize $\delta_1^\text{p}$\tiny$\uparrow$ & \scriptsize Rel\textsuperscript{p}\tiny$\downarrow$ & \scriptsize $\delta_1^\text{p}$\tiny$\uparrow$ & \scriptsize Rel\textsuperscript{p}\tiny$\downarrow$ & \scriptsize $\delta_1^\text{p}$\tiny$\uparrow$ & \scriptsize Rel\textsuperscript{p}\tiny$\downarrow$ & \scriptsize $\delta_1^\text{p}$\tiny$\uparrow$ & \scriptsize Rank\tiny$\downarrow$\\
        \hline
        \multicolumn{19}{c}{Scale-invariant point map}\\
        \hline
        
        LeReS & 16.9 & 76.0 & 31.6 & 28.4 & 17.1 & 75.8 & 18.5 & 72.2 & 14.7 & 76.0 & 38.6 & 30.6 & 32.0 & 39.4 & 27.6 & 46.4 & 24.6 & 55.6 & 3.94 \\
        
        DUSt3R & 5.53 & 97.1 & 15.2 & 87.9 & \underline{10.7} & \underline{90.6} & 6.18 & 95.4 & \underline{4.54} & 99.3 & 34.8 & \underline{50.3} & 21.4 & 70.1 & 12.4 & 86.7 & 13.8 & 84.7 & 2.75 \\
        
        UniDepth & \underline{5.33} & \textbf{98.4} & \underline{5.96} & \textbf{98.5} & 18.5 & 77.6 & \underline{5.29} & \textbf{97.4} & 6.58 & \underline{99.6} & \underline{33.0} & 48.9 & \textbf{11.4} & \underline{90.2} & \underline{12.3} & \underline{91.0} & \underline{12.3} & \underline{87.7} & \underline{2.09} \\
        
        \textbf{Ours} & \textbf{4.86} & \textbf{98.4} & \textbf{5.47} & \underline{97.4} & \textbf{4.58} & \textbf{98.9} & \textbf{4.63} & \underline{97.1} & \textbf{2.58} & \textbf{100} & \textbf{22.3} & \textbf{69.5} & \underline{12.3} & \textbf{90.3} & \textbf{6.58} & \textbf{94.5} & \textbf{7.91} & \textbf{93.3} & \textbf{1.22} \\
        
        \hline  
        \multicolumn{19}{c}{Affine-invariant point map}\\
        \hline
        
        LeReS & 9.51 & 91.4 & 26.1 & 49.1 & 14.7 & 79.6 & 11.0 & 88.6 & 8.91 & 95.2 & 29.7 & 55.5 & 29.4 & 46.7 & 15.1 & 80.1 & 18.1 & 73.3 & 3.94 \\
        
        DUSt3R & 4.45 & 97.4 & 12.7 & 83.3 & \underline{7.27} & \underline{95.0} & 5.04 & 96.0 & 3.07 & 99.6 & 30.3 & 56.6 & 19.7 & 71.2 & 8.97 & 88.7 & 11.4 & 86.0 & 2.94 \\
        
        UniDepth & \underline{3.93} & \textbf{98.4} & \textbf{4.29} & \textbf{98.6} & 12.2 & 89.6 & \underline{4.65} & \textbf{98.0} & \underline{2.99} & \underline{99.8} & \underline{28.5} & \underline{58.4} & \textbf{10.3} & \underline{90.5} & \underline{8.56} & \underline{90.9} & \underline{9.43} & \underline{90.5} & \underline{1.81} \\
        
        \textbf{Ours} & \textbf{3.68} & \underline{98.3} & \underline{4.86} & \underline{97.2} & \textbf{3.57} & \textbf{99.0} & \textbf{3.61} & \underline{97.3} & \textbf{1.14} & \textbf{100} & \textbf{16.8} & \textbf{77.8} & \underline{10.5} & \textbf{91.4} & \underline{4.37} & \textbf{96.4} & \textbf{6.07} & \textbf{94.7} & \textbf{1.31} \\
        
        \hline
        \multicolumn{19}{c}{Local point map}\\
        \hline
        
        LeReS & - & - & - & - & 9.32 & 91.9 & 8.57 & 93.2 & - & - & 13.3 & 84.8 & 10.7 & 88.9 & 11.6 & 88.2 & 10.7 & 89.4 & 3.80 \\
        
        DUSt3R & - & - & - & - & \underline{6.05} & \underline{94.8} & \underline{5.44} & 95.9 & - & - & \underline{11.8} & \underline{87.0} & 9.24 & 90.8 & \underline{7.32} & \underline{93.1} & \underline{7.97} & \underline{92.3} & \underline{2.30} \\
        
        UniDepth & - & - & - & - & 8.61 & 92.6 & 5.92 & \underline{96.0} & - & - & 13.4 & 84.3 & \underline{8.18} & \underline{92.0} & 9.95 & 90.0 & 9.21 & 91.0 & 2.90 \\
        
        \textbf{Ours} & - & - & - & - & \textbf{3.21} & \textbf{98.1} & \textbf{4.16} & \textbf{96.8} & - & - & \textbf{8.63} & \textbf{92.7} & \textbf{6.74} & \textbf{94.3} & \textbf{4.78} & \textbf{96.3} & \textbf{5.50} & \textbf{95.6} & \textbf{1.00} \\
        
        \specialrule{.12em}{0.12em}{0em}
    \end{tabular}
    \vspace{-5pt}
    \caption{
    Quantitative results for point map estimation. Rel$^\text{p}$ and $\delta_1^\text{p}$ are in percentage. 
    The best values are highlighted in \textbf{bold}, and the second-best ones are \underline{underlined}. Local point map accuracy is evaluated on affine-invariant point maps within local object regions
    }
    \label{tab:comparison_points}
    \vspace{-5pt}
\end{table*}

\begin{table*}[h]
    \footnotesize
    \centering
    \setlength{\tabcolsep}{4pt}
    \begin{tabular}{l|cc|cc|cc|cc|cc|cc|cc|cc|ccc}
        \specialrule{.12em}{0em}{0em}
        \multirow{2}{*}{\textbf{Method}} & \multicolumn{2}{c|}{NYUv2} & \multicolumn{2}{c|}{KITTI} & \multicolumn{2}{c|}{ETH3D} & \multicolumn{2}{c|}{iBims-1} & \multicolumn{2}{c|}{GSO} & \multicolumn{2}{c|}{Sintel} & \multicolumn{2}{c|}{DDAD} & \multicolumn{2}{c|}{DIODE} & \multicolumn{3}{c}{Average}\\
        & \scriptsize Rel\textsuperscript{d}\tiny$\downarrow$ & \scriptsize $\delta_1^\text{d}$\tiny$\uparrow$ & \scriptsize Rel.\tiny$\downarrow$ & \scriptsize $\delta_1^\text{d}$\tiny$\uparrow$ & \scriptsize Rel\textsuperscript{d}\tiny$\downarrow$ & \scriptsize $\delta_1^\text{d}$\tiny$\uparrow$ & \scriptsize Rel.\tiny$\downarrow$ & \scriptsize $\delta_1^\text{d}$\tiny$\uparrow$ & \scriptsize Rel\textsuperscript{d}\tiny$\downarrow$ & \scriptsize $\delta_1^\text{d}$\tiny$\uparrow$ & \scriptsize Rel\textsuperscript{d}\tiny$\downarrow$ & \scriptsize $\delta_1^\text{d}$\tiny$\uparrow$ & \scriptsize Rel\textsuperscript{d}\tiny$\downarrow$ & \scriptsize $\delta_1^\text{d}$\tiny$\uparrow$ & \scriptsize Rel\textsuperscript{d}\tiny$\downarrow$ & \scriptsize $\delta_1^\text{d}$\tiny$\uparrow$ & \scriptsize Rel\textsuperscript{d}\tiny$\downarrow$ & \scriptsize $\delta_1^\text{d}$\tiny$\uparrow$ & \scriptsize Rank\tiny$\downarrow$\\
        
        \hline
        \multicolumn{19}{c}{Scale-invariant depth map}\\
        \hline
        
LeReS & 12.1 & 82.6 & 19.2 & 64.8 & 14.2 & 78.4 & 14.0 & 78.8 & 13.6 & 77.9 & 30.5 & 52.1 & 26.5 & 52.0 & 18.2 & 69.6 & 18.5 & 69.5 & 7.31 \\

ZoeDepth & \textcolor{lightgray}{5.62} & \textcolor{lightgray}{96.3} & \textcolor{lightgray}{7.27} & \textcolor{lightgray}{91.9} & 10.4 & 87.3 & 7.45 & 93.2 & 3.23 & \underline{99.9} & 27.4 & 61.8 & 17.0 & 72.8 & 11.3 & 85.2 & \textcolor{lightgray}{11.2} & \textcolor{lightgray}{86.1} & 5.50 \\

DUSt3R & 4.40 & 97.1 & 7.81 & 90.6 & 6.04 & 95.7 & 4.98 & 95.8 & 3.27 & 99.5 & 31.1 & 57.2 & 18.6 & 73.3 & 8.91 & 88.8 & 10.6 & 87.2 & 5.00 \\

Metric3D V2 & 4.69 & 97.4 & \underline{4.00} & \underline{98.5} & \underline{3.84} & \underline{98.5} & \underline{4.23} & \underline{97.7} & \underline{2.46} & \underline{99.9} & \underline{20.7} & \underline{69.8} & \textcolor{lightgray}{7.41} & \textcolor{lightgray}{94.6} & \textbf{3.29} & \textbf{98.4} & \textcolor{lightgray}{6.33} & \textcolor{lightgray}{94.3} & \underline{2.07} \\

UniDepth & \underline{3.86} & \textbf{98.4} & \textbf{3.73} & \textbf{98.6} & 5.67 & 97.0 & 4.79 & 97.4 & 4.18 & 99.7 & 28.3 & 58.8 & \underline{10.1} & \textbf{90.5} & 6.83 & 92.8 & \underline{8.43} & \underline{91.6} & {3.00} \\

DA V1 & \textcolor{lightgray}{4.77} & \textcolor{lightgray}{97.5} & \textcolor{lightgray}{5.61} & \textcolor{lightgray}{95.6} & 9.41 & 88.9 & 5.53 & 95.8 & 5.49 & 99.3 & 28.3 & 56.7 & 13.2 & 81.5 & 10.3 & 87.5 & \textcolor{lightgray}{10.3} & \textcolor{lightgray}{87.9} & 5.67 \\

DA V2 & 5.03 & 97.3 & 7.23 & 93.7 & 6.12 & 95.5 & 4.32 & \textbf{97.9} & 4.38 & 99.3 & 23.0 & 65.2 & 14.7 & 78.0 & 7.95 & 90.0 & 9.09 & 89.6 & 4.06 \\

\textbf{Ours} & \textbf{3.44} & \textbf{98.4} & 4.25 & 97.8 & \textbf{3.36} & \textbf{98.9} & \textbf{3.46} & 97.0 & \textbf{1.47} & \textbf{100} & \textbf{19.3} & \textbf{73.4} & \textbf{9.17} & \textbf{90.5} & \underline{4.89} & \underline{94.7} & \textbf{6.17} & \textbf{93.8} & \textbf{1.62} \\
        
        \hline
        \multicolumn{19}{c}{Affine-invariant depth map}\\
        \hline
        
        Marigold  & \underline{4.63} & 97.3 & \underline{7.29} & \underline{93.8} & \underline{6.08} & \underline{96.3} & \underline{4.35} & \underline{97.2} & 2.78 & \underline{99.9} & 21.2 & 75.0 & \underline{14.6} & \underline{80.5} & \underline{6.34} & \underline{94.3} &\underline{8.41}&\underline{91.8} &\underline{2.25}\\
        
        GeoWizard & 4.69 & \underline{97.4} & 8.14 & 92.5 & 6.90 & 94.0 & 4.50 & 97.1 & \underline{2.00} & \underline{99.9} & \underline{17.8} & \underline{76.2} & 16.5 & 75.7 & 7.03 & 92.7 &8.44&90.7 &2.69\\ 
                
        \textbf{Ours}  & \textbf{2.92} & \textbf{98.6} & \textbf{3.94} & \textbf{98.0} & \textbf{2.69} & \textbf{99.2} & \textbf{2.74} & \textbf{97.9} & \textbf{0.94} & \textbf{100} & \textbf{13.0} & \textbf{83.2} & \textbf{8.40} & \textbf{92.1} & \textbf{3.16} & \textbf{97.5} & \textbf{4.72} & \textbf{95.8} & \textbf{1.00}\\
        
        \hline  
        \multicolumn{19}{c}{Affine-invariant disparity map}\\
        \hline
        
MiDaS V3.1 & 4.58 & 98.1 & 6.25 & 94.7 & 5.77 & 96.8 & 4.73 & 97.4 & 1.86 & \textbf{100} & 21.3 & 73.1 & 14.5 & 82.6 & 6.05 & 94.9 & 8.13 & 92.2 & 3.69 \\

DA V1 & 4.20 & \underline{98.4} & \underline{5.40} & \underline{97.0} & \underline{4.68} & \underline{98.2} & 4.18 & 97.6 & 1.54 & \textbf{100} & \underline{20.1} & \underline{77.6} & \underline{12.7} & \underline{86.9} & 5.69 & 95.7 & \underline{7.31} & \underline{93.9} & \underline{2.31} \\

DA V2 & \underline{4.14} & 98.3 & 5.61 & 96.7 & 4.71 & 97.9 & \underline{3.47} & \textbf{98.5} & \underline{1.24} & \textbf{100} & 21.4 & 72.8 & 13.1 & 86.4 & \underline{5.29} & \underline{96.1} & 7.37 & 93.3 & 2.56 \\

\textbf{Ours} & \textbf{3.38} & \textbf{98.6} & \textbf{4.05} & \textbf{98.1} & \textbf{3.11} & \textbf{98.9} & \textbf{3.23} & \underline{98.0} & \textbf{0.96} & \textbf{100} & \textbf{18.4} & \textbf{79.5} & \textbf{8.99} & \textbf{91.5} & \textbf{3.98} & \textbf{97.2} & \textbf{5.76} & \textbf{95.2} & \textbf{1.06} \\
        
        \specialrule{.12em}{0.12em}{0em}
    \end{tabular}
    \vspace{-5pt}
    \caption{Quantitative results for depth map estimation. \textcolor{lightgray}{Gray numbers} denote models trained on respective benchmarks and thus excluded from ranking. A more extensive comparison can be found in the \emph{suppl. materials}.}
    \label{tab:comparison_depth}
    \vspace{-5pt}
\end{table*}

\begin{table*}[ht!]  
\centering  
\begin{minipage}[t]{0.38\linewidth}  
    \footnotesize
    \setlength{\tabcolsep}{1.5pt}
    \centering
    \begin{tabular}{l|cc|cc|cc|ccc}
        \specialrule{.12em}{0em}{0em}
           \multirow{2}{*}{\textbf{Method}} & \multicolumn{2}{c|}{NYUv2}  &\multicolumn{2}{c|}{ETH3D} & \multicolumn{2}{c|}{iBims-1} & \multicolumn{3}{c}{Average}\\
        &\tiny Mean\tiny$\downarrow$&\tiny Med.\tiny$\downarrow$ &\tiny Mean&\tiny Med.\tiny$\downarrow$&\tiny Mean\tiny$\downarrow$&\tiny Med.\tiny$\downarrow$ &\tiny Mean\tiny$\downarrow$&\tiny Med.\tiny$\downarrow$&\tiny Rank$\downarrow$\\
        \hline
        Perspective &5.38&4.39 &13.6&11.9 &10.6&9.30 &9.86&8.53 &5.00\\
        
        WildCam &{3.82}&\underline{3.20} &7.70&5.81 &9.48&9.08  &7.00&6.03 &3.00\\
        
        \hline
        LeReS &19.4&19.6 &8.26&7.19 &18.4&17.5 &15.4&14.8&5.53\\
        
        DUSt3R &\textbf{2.57}&\textbf{1.86} &\underline{5.77}&\underline{3.60} &\underline{3.83}&\underline{2.53}  &\underline{4.06}&\underline{2.66} &\underline{1.67}\\
        
        UniDepth &7.56&4.31 &10.7&9.96 &11.9&5.96  &10.1&6.74 &4.50\\
        
        \textbf{Ours} &\underline{3.41}&\underline{3.21} &\textbf{2.50}&\textbf{1.54} &\textbf{2.81}&\textbf{1.89}  &\textbf{2.91}&\textbf{2.21} &\textbf{1.50}\\
        \specialrule{.12em}{0em}{0em}
    \end{tabular}
    \vspace{-5pt}
    \caption{Evaluation results for camera FOV in degrees. }
    \label{tab:comparison_fov}
\end{minipage}  
\hfill  
\begin{minipage}[t]{0.58\linewidth}  
    \scriptsize
    \setlength{\tabcolsep}{2.5pt}
    \centering
    \begin{tabular}{l|cc|cc|cc|cc|cc|cc}
        \specialrule{.12em}{0em}{0em}
        \multirow{3}{*}{\textbf{Ablation}} & \multicolumn{6}{c|}{Point} & \multicolumn{4}{c|}{Depth} & \multicolumn{2}{c}{Disparity} \\
        
        & \multicolumn{2}{c|}{\scriptsize Scale-inv.} & \multicolumn{2}{c|}{\scriptsize Affine-inv.} & \multicolumn{2}{c|}{\scriptsize Local} & \multicolumn{2}{c|}{\scriptsize Scale-inv.} & \multicolumn{2}{c|}{\scriptsize Affine-inv.} & \multicolumn{2}{c}{\scriptsize Affine-inv.}\\ 
        
        &\tiny Rel\textsuperscript{p}\tiny$\downarrow$
        &\tiny $\delta_1^\text{p}$\tiny$\uparrow$ 
        &\tiny Rel\textsuperscript{p}\tiny$\downarrow$
        &\tiny $\delta_1^\text{p}$\tiny$\uparrow$ 
        &\tiny Rel\textsuperscript{p}\tiny$\downarrow$
        &\tiny $\delta_1^\text{p}$\tiny$\uparrow$ 
        &\tiny Rel\textsuperscript{d}\tiny$\downarrow$
        &\tiny $\delta_1^\text{d}$\tiny$\uparrow$ 
        &\tiny Rel\textsuperscript{d}\tiny$\downarrow$
        &\tiny $\delta_1^\text{d}$\tiny$\uparrow$ 
        &\tiny Rel\textsuperscript{d}\tiny$\downarrow$
        &\tiny $\delta_1^\text{d}$\tiny$\uparrow$  \\
        \hline
        
SI-Log depth & 11.2 & 88.7 & 9.09 & 90.6 & 9.19 & 91.2 & 8.94 & 90.1 & 7.27 & 92.6 & 8.23 & 92.1 \\

Affine-inv. depth & 29.9 & 51.4 & 29.0 & 52.7 & 12.2 & 86.0 & 28.9 & 52.7 & \textbf{6.18} & \textbf{93.9} & 15.9 & 76.6 \\

ROE scale-inv. & 10.3 & 89.8 & 8.34 & 91.6 & 8.59 & 91.9 & 8.27 & 90.9 & 6.73 & 93.2 & 7.90 & 92.6 \\

L2 affine-inv. & 13.5 & 84.2 & 10.3 & 88.2 & 9.48 & 91.0 & 11.1 & 85.7 & 8.03 & 91.2 & 9.37 & 90.5 \\

Med. affine-inv. & 10.9 & 89.0 & 8.97 & 90.7 & 9.44 & 90.7 & 9.10 & 89.8 & 7.50 & 92.4 & 8.74 & 91.8 \\

\textbf{ROE affine-inv.} & \textbf{9.84} & \textbf{90.3} & \textbf{7.88} & \textbf{92.1} & \textbf{7.62} & \textbf{93.3} & \textbf{7.91} & \textbf{91.2} & 6.29 & 93.7 & \textbf{7.43} & \textbf{93.2} \\
        
        \hline

\textbf{Full} \textit{{w/o}} trunc. & 9.81 & 90.5 & 7.91 & 91.7 & \textbf{7.12} & \textbf{93.8} & 7.92 & \textbf{91.3} & 6.31 & 93.5 & 7.45 & 93.1 \\

\textbf{Full} \textbf{w/o} $\mathcal L_S$ & 9.98 & 90.3 & 7.94 & \textbf{92.1} & 7.47 & 93.4 & 7.94 & 91.2 & 6.30 & 93.6 & 7.47 & 93.2 \\

\textbf{Full} & \textbf{9.78} & \textbf{90.6} & \textbf{7.83} & \textbf{92.1} & 7.16 & \textbf{93.8} & \textbf{7.82} & \textbf{91.3} & \textbf{6.20} & \textbf{93.7} & \textbf{7.30} & \textbf{93.3} \\
        
        \specialrule{.12em}{0em}{0em}
    \end{tabular}
    \vspace{-5pt}
    \caption{Quantitative ablation study results. All experiments are conducted with a ViT-Base encoder. The first six rows are trained with global loss only.}
    \label{tab:ablation}
\end{minipage}  
\vspace{-10pt}
\end{table*}

\begin{figure*}[ht!]
    \centering
    \includegraphics[width=\linewidth]{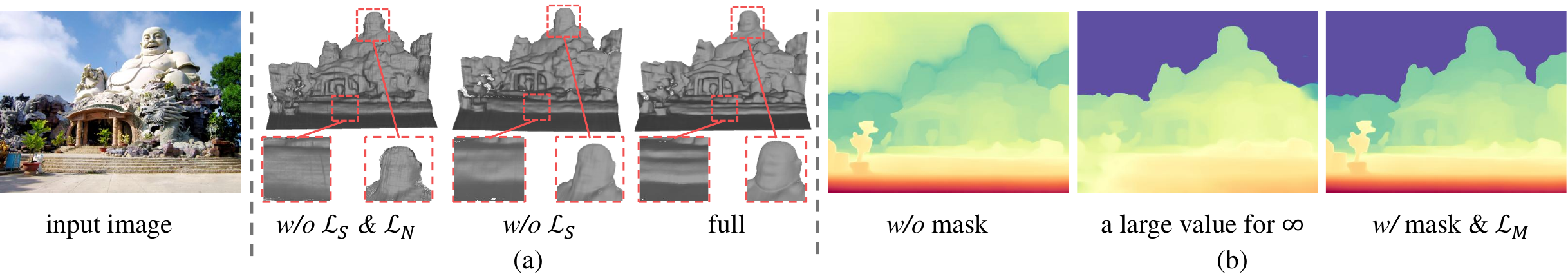}
    \vspace{-15pt}
    \caption{
        Qualitative ablation study results. All experiments use the ViT-Base backbone for the encoder.
        \textbf{(a)} Surface visualization of the impact of $\mathcal{L}_S$ and $\mathcal{L}_N$. Removing either leads to noisy surfaces and poor geometry. 
        \textbf{(b)} Depth visualization for the ablation of valid region mask prediction. Our method correctly predicts the sky regions, for which the predicted point values will be erroneous if the masks are removed. Supervising infinity regions by assigning a large distance label can negatively affect the foreground prediction accuracy.
    }
    \label{fig:ablation}
    \vspace{-10pt}
\end{figure*}

\paragraph{Implementation details.}
We use a ViT~\cite{dosovitskiy2020vit} encoder pretrained with DINOv2~\cite{oquab2023dinov2} and a lightweight CNN-based upsampler as the decoder.
The encoder and decoder have initial learning rates of $5\times 10^{-6}$ and $5\times 10^{-5}$, with a decay factor of 5 every 100K iterations. The training batch size is $256$.
Our model is trained on images with aspect ratios uniformly sampled between $1:2$ and $2:1$, and pixel counts from 250K to 500K. We apply several image augmentations, including color jittering, Gaussian blurring, JPEG compression-decompression, and random cropping. The principal point of the cropped image is aligned with its center using a perspective crop.

\subsection{Quantitative Evaluations. }
We assess the zero-shot performance of our model and compare it to several state-of-the-art methods for monocular point map estimation~\cite{wei2021leres, piccinelli2024unidepth, wang2024dust3r} and depth map estimation~\cite{yang2024depthanything, yang2024depthanythingv2, birkl2023midas31, hu2024metric3dv2, bhat2023zoedepth, fu2024geowizard, ke2024marigold}. To ensure a fair comparison, all methods utilize ViT-Large~\cite{dosovitskiy2020vit} as backbone, except for LeReS (using ResNeXt101~\cite{xie2017aggregated}) as well as GeoWizard and Marigold (using StableDiffusion V1~\cite{rombach2021ldmSD}).
The scale and shift of the predicted point maps and depth maps are aligned with the ground truth when necessary. 
We also evaluate the FOV prediction results against several recent methods~\cite{jin2023perspectivefields, zhu2024wildcamera}. As detailed below, \emph{our method achieves superior results across various tasks and benchmarks}.\footnote{In this revision, we corrected evaluation numbers due to a minor error in the evaluation data. The overall rankings and conclusions are not affected. All results are reproducible using the released code.}

\paravspace
\paragraph{Point map estimation.} 
We evaluate the accuracy of point map estimation on eight diverse datasets: NYUv2~\cite{Silberman2012nyuv2}, KITTI~\cite{Uhrig2017kitti}, ETH3D~\cite{Schops2019ETH3D}, iBims-1~\cite{ibim1_1}, Sintel~\cite{Butler2012sintel}, Google Scanned Objects (GSO)~\cite{downs2022googlescannedobjects}, DDAD~\cite{ddadpacking}, and DIODE~\cite{diode_dataset}. These datasets represent a wide range of domains, including indoor scenes, street views, object scans, and synthetic movies. The raw datasets are processed for reliable evaluation (\eg, sky regions in Sintel dataset and boundary artifacts in the DIODE are removed; see the \emph{suppl. material} for more preprocessing details). For evaluation metrics, we use the relative point error Rel\textsuperscript{p}, \ie, $\|\hat{\mathbf{p}} - \mathbf{p}\|_2 / \|\mathbf{p}\|_2$, and the percentage of inliers $\delta_1^\text{p}$ with $\|\hat{\mathbf{p}} - \mathbf{p}\|_2 / \min(\|\mathbf{p}\|_2,\|\hat {\mathbf{p}}\|_2) < 0.25$.

We compare our method with LeReS~\cite{wei2021leres}, UniDepth~\cite{piccinelli2024unidepth}, and DUSt3R~\cite{wang2024dust3r}, evaluating both camera-space scale-invariant point maps and affine-invariant point maps. Table~\ref{tab:comparison_points} demonstrates that our method outperforms existing methods, achieving the lowest average Rel\textsuperscript{p} and highest $\delta_1^\text{p}$ for both point map representations.

To further assess local geometry accuracy, we evaluate affine-invariant point maps within regions defined by object segmentation masks, sourced from dataset labels or Segment Anything~\cite{kirillov2023sam}.
Due to the need for high-quality ground truth and scenes with multiple objects, NYUv2, KITTI, and GSO are excluded.
Our method achieves a substantial reduction in region-wise Rel\textsuperscript{p}, lowering the error from $7.97$ to $5.50$—an approximately $30\%$ improvement over previous methods.

\paravspace
\paragraph{Depth map estimation.} 
We employ the same benchmarks used for point map evaluation to assess the depth estimation accuracy. We adopt absolute relative error Rel\textsuperscript{d}, \ie, $|z-\hat z|/z$, and percentage of inliers $\delta_1^d$ with $\max(z/\hat z,\hat z/z)<1.25$ as the evaluation metrics.

For a comprehensive comparison with different monocular depth estimation methods, we evaluate scale-invariant depth~\cite{wei2021leres, bhat2023zoedepth, wang2024dust3r, hu2024metric3dv2, piccinelli2024unidepth}, affine-invariant depth~\cite{ke2024marigold, fu2024geowizard}, and affine-invariant disparity~\cite{birkl2023midas31, yang2024depthanything, yang2024depthanythingv2}. 
The $z$ coordinates in our predicted point maps represent affine-invariant depth, which can be converted to scale-invariant depth using the shift derived from Eq.~\ref{eq:reover_shift_focal}.
For affine-invariant disparity, we take the inverse of our scale-invariant depth predictions.
As shown in Table~\ref{tab:comparison_depth}, our method exhibits the lowest average Rel\textsuperscript{d} for all depth representations. Since existing methods with scale-invariant depth representations can be converted to affine-invariant depth and disparity, we provide an extensive comparison in the \emph{suppl. materials}, where our method also significantly outperforms the others.

\paravspace
\paragraph{Camera FOV estimation.} 
To evaluate the accuracy of the estimated camera FOV, we select benchmarks with $\text{vertical FOV}\geq 45^\circ$, including NYUv2~\cite{Silberman2012nyuv2}, ETH3D~\cite{Schops2019ETH3D} and iBims-1~\cite{ibim1_1}, and augment the FOV by randomly center cropping 50\%$\sim$100\% from the original images. 
All input images are undistorted with a centered principal point.

We compare our model to point map estimation methods \cite{wei2021leres,piccinelli2024unidepth,wang2024dust3r} as well as two learning-based camera calibration methods: Perspective Fields~\cite{jin2023perspectivefields} and Wild Camera~\cite{zhu2024wildcamera}. 
For fair zero-shot evaluation, we utilize their pre-trained checkpoints on in-the-wild datasets. Table~\ref{tab:comparison_fov} reports mean and median FOV errors. Our method achieves a mean error of $2.91^\circ$ and a median error of $2.21^\circ$ on average of the three FOV benchmarks, outperforming both camera calibration and point map estimation methods.
 
\subsection{Qualitative comparison} 
\begin{figure*}[ht!]
    \centering
    \includegraphics[width=\linewidth]{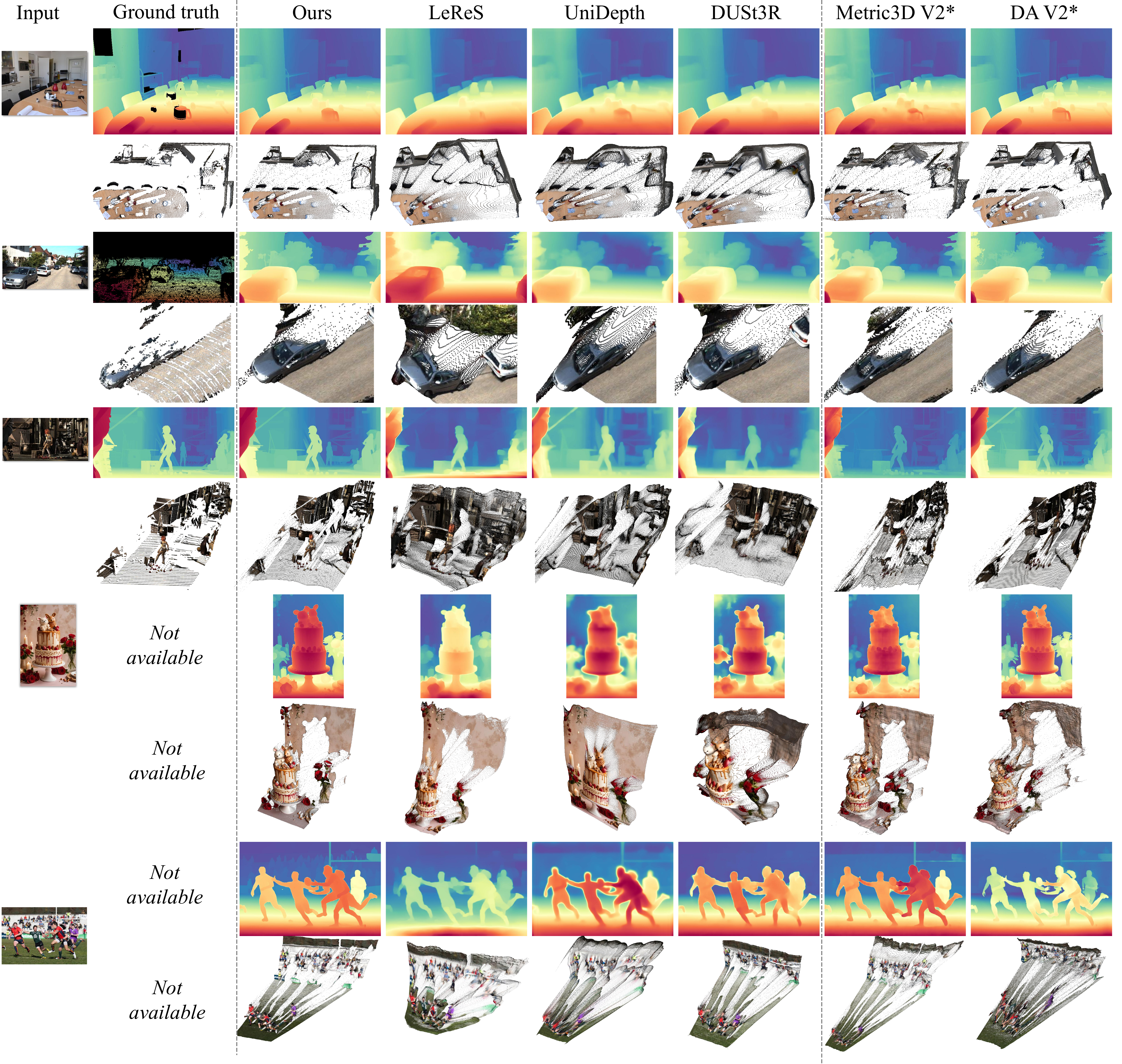}
    \vspace{-10pt}
    \caption{Qualitative comparison of point and disparity maps. The first three cases are from our test datasets, while the last two rows are random ``in-the-wild" images. *: for methods without camera intrinsics prediction, we use ground-truth camera intrinsics (and disparity shifts) to lift their results into 3D points when available, or use our intrinsics prediction for the  ``in-the-wild" images. \textbf{\emph{Best view with zoom.}}}
    \label{fig:qualitative_comparison}
    \vspace{-15pt}
\end{figure*}

We show visual results of disparity and point maps of different methods in Fig.~\ref{fig:qualitative_comparison}. 
For depth prediction methods without camera parameters estimation, we lift their results to 3D points using ground-truth intrinsics when available. Upon visual inspection, our method produces significantly less-distorted point maps compared to others, demonstrating its superior generalizability and accuracy. For example, as shown in the third row of Fig.~\ref{fig:qualitative_comparison}, other methods show noticeable stretching and deformation of the car shape. Although \cite{yang2024depthanythingv2} produces visually sharp disparity maps, distorted 3D geometry is evident even with ground-truth camera intrinsics. 
In contrast, our method yields a more regular and well-formed car structure. More comparisons can be found in the \emph{suppl. materials}.

\subsection{Ablation Study}
We conducted comprehensive ablation studies on our point map representation, alignment strategy and loss function designs. 
ViT-base backbones are used here for efficiency.

\paravspace
\paragraph{Point map representation.}
We compare our affine-invariant point map representation with three variants: (1) a scale-invariant depth map and a dense ray map for camera parameters, (2) an affine-invariant depth map with a shift scalar and a dense ray map, and (3) a scale-invariant point map in camera space. The first variant uses SI-log loss~\cite{eigen2014depth} for the depth map and MSE for the ray map, follows UniDepth~\cite{piccinelli2024unidepth}. The second employs ROE alignment with $L_1$ loss for the affine depth, $L_1$ loss for shift, and MSE for the ray map. The third applies the ROE alignment to solve the scale parameter only.
Table~\ref{tab:ablation} (row 1-3) shows that our affine-invariant representation outperforms the others across various tasks. Notably, while the affine-invariant depth representation slightly improves specific metrics, all others drop dramatically. 
Our representation excels by effectively resolving focal-distance ambiguity, a key issue that hampers effectively training of the other three variants.

\paravspace
\paragraph{ROE alignment.}
To assess the effectiveness of our optimal alignment, we compare it to least-squares alignment for $L_2$ loss and a previously-used median-based alignment~\cite{ranftl2020midas} for $L_1$ loss.  The former has an analytic solution, while the latter normalizes predicted and ground-truth point maps by their median: $(\mathbf p - \mathbf{t}(\mathbf{P}))/s(\mathbf{P})$, where $\mathbf t(\mathbf P)=(0, 0,\operatorname{median}(\mathbf Z))$, and $s(\mathbf P)=\frac{1}{N}\sum_{i=1}^N\|\mathbf p_i-\mathbf t(\mathbf P)\|_1$. As shown in the Table~\ref{tab:ablation} (row 4-6), our proposed alignment method consistently outperforms the other two strategies across all tasks by large margins, highlighting its critical role. We also ablate the truncation of alignment objective residuals, which degrades performance (Table~\ref{tab:ablation} row 7).

\paravspace
\paragraph{Multi-scale local geometry loss.}
As shown in Table~\ref{tab:ablation} (row 8), the quantitative results without $\mathcal L_S$ suffer a significant drop on local geometry metrics. This demonstrates the effectiveness of $\mathcal L_S$ in enhancing local geometry accuracy and compensating for the insufficient supervision of the global loss due to the ambiguous relative positions between distant scene objects. Fig.~\ref{fig:ablation}~(a) further demonstrates the effectiveness of the proposed local geometry loss.

\paravspace
\paragraph{Infinity mask.} 
Our method predicts a valid region mask to handle infinity regions. 
As shown in Fig.~\ref{fig:ablation}~(b), it correctly identifies sky regions, where the output point values would be erroneous if the masks were removed.
Another strategy for addressing infinity is to assign a large value, \ie, $1,000\times$ the average distance, to these regions. Fig.~\ref{fig:ablation}~(b) shows that this strategy compromises the foreground accuracy as the network struggles with large values for infinity. 
Incorporating a separate mask prediction effectively handles infinity while retaining accurate foreground geometry.

\section{Conclusion}

We have presented a method for accurate monocular geometry estimation of open-domain images. Our key insight is to design effective supervision which is largely neglected by previous methods. We introduce an affine-invariant point map representation which precludes ambiguous supervision during training. A robust, optimal, and efficient point map alignment solver is introduced for effective shape learning, along with a multi-scale geometry loss for precise local supervision. Trained on a large dataset collection, our model demonstrates strong generalizability and significantly outperforms previous methods across various tasks and benchmarks. We believe our method marks a significant advancement in monocular 3D geometry estimation and can serve as a robust foundational model for various applications.

{
    \small
    \bibliographystyle{ieeenat_fullname}
    \bibliography{main}
}

\clearpage
\setcounter{page}{1}
\maketitlesupplementary

\appendix

\section{Algorithm Details}

\subsection{Recovering Shift and Camera Focal}

We assume a simple pinhole camera model with isotropic focal length and centered principal point. 
The 2D image plane is parameterized with the center as $(0, 0)$. 
The image plane coordinate of pixel $i$ is denoted as $(u_i, v_i)$, corresponding to its predicted 3D point $\mathbf p_i=(x_i, y_i, z_i)$. The focal length and shift is obtained by minimizing the projection error,
\begin{equation}
    \min_{f,t_z} \sum_{i\in\mathcal M} \left({fx_i\over z_i + t_z'} - u_i \right)^2 + \left({fy_i\over z_i + t_z'} - v_i \right)^2,
    \label{eq:reover_shift_focal_}
\end{equation}
which can be further reduced to have a single variable $t_z'$ by substituting $f$ with its close-form solution with respect to $t_z'$, 
\begin{equation}
    f={\sum_{i\in \mathcal M} \left(x_i\over z_i+t_z'\right)u_i+\sum_{i\in \mathcal M} \left(y_i\over z_i+t_z'\right)v_i\over \sum_{i\in \mathcal M} \left(x_i\over z_i+t_z'\right)^2 + \sum_{i\in \mathcal M}\left(y_i\over z_i+t_z'\right)^2}.
\end{equation}
We use a numerical solver for this least squares problem with Levenberg-Marquardt algorithm~\cite{more2006levenberg} implemented by SciPy~\cite{2020SciPy-NMeth} package. For efficiency, the point map is resized to low resolution ($64\times64$) for running this algorithm. In our practice, it typically converges within 10 iterations in around 3ms.

\subsection{ROE Alignment}\label{sec:supp_opt_align}

We will first introduce an algorithm to a simpler subproblem then derive the solution to either with or without the constraint of $t_x=t_y=0$ (1D-shift case or 3D-shift case, respectively). 

\paravspace
\paragraph{Subproblem (\textit{w/o} truncation).}

Consider the optimization objective with respect to scale $s$ only, denoted as $l_0(s)$. We omit the mask $\mathcal M$ for simplicity and denote $N$ as the number of valid points:
\begin{equation}
    \min_s l_0(s) = \min_s \sum_{i=1}^N w_i|s\hat x_i-x_i|,
\label{eq:subproblem_wo_truncate}
\end{equation}
where $w_i>0$ and $\hat x_i> 0$ without loss of generality. The objective, as a summation of convex functions, is also convex obviously. The minimum occurs where its left-hand $l_0'^-(s)$ derivative and right-hand derivative $l_0'^+(s)$ have opposite signs or one of them is zero,  
\begin{equation}
\begin{split}
    l_0'^{-}(s)
    &=\!\!\sum_{{x_i\over\hat  x_i}<s}{w_i\hat x_i}-\!\!\sum_{s\leq{x_i\over\hat  x_i}}{w_i\hat x_i}
    =2\sum_{{x_i\over\hat  x_i}<s}w_i\hat x_i-\!\!\sum_{i=1}^N w_i\hat x_i,
    \\
    l_0'^{+}(s)
    &=\!\!\sum_{{x_i\over\hat  x_i}\leq s}{w_i\hat x_i}-\!\!\sum_{s<{x_i\over\hat  x_i}}{w_i\hat x_i}
    =2\sum_{{x_i\over\hat  x_i}\leq s}w_i\hat x_i-\!\!\sum_{i=1}^N w_i\hat x_i.
\end{split}
\end{equation}
$l_0'^-(s)$ and $l_0'^+(s)$ differ at $\{{x_i\over\hat  x_i}\}$. First,
we sort $\{{x_i\over\hat x_i}\}_{i=1}^N$ and compute the prefix summations of $\{w_i\hat x_i\}_{i=1}^N$. This allows us to evaluate the derivatives in $O(1)$ time for each point in $\{{x_i\over\hat  x_i}\}_{i=1}^N$. Finally, $\hat x_i\over x_i$ such that $\l_0'^-({x_i\over\hat  x_i})\leq0\leq\l_0'^+({x_i\over\hat  x_i})$ is the minimum point. The solution is outlined in Algorithm~\ref{alg:subproblem_wo_truncation}.

\begin{figure}[t]
\vspace{-10pt}
\begin{algorithm}[H]
\caption{ROE alignment subproblem \textit{w/o} truncation}
\label{alg:subproblem_wo_truncation}
\begin{algorithmic}    
\State \textbf{input:} arrays $\hat X[1..n]$, $X[1..n]$, $W[1..n]$
\State \textbf{output:} optimal scale $s^*$ and objective value $l^*$ to Eq.~\ref{eq:subproblem_wo_truncate}
\Statex
\Function{SolveSubproblem}{$\hat X$, $X$, $W$}
	\State sort arrays $\hat X$, $X$, $W$ by {\small $X[i]/\hat X[i]$}
	\State $Q[1..n]$ $\gets$ accumulated sum of {\small $W*\hat X$}
    \State $D[0..n]$ $\gets$ {\small  $\{-Q[n]\} \cup \{2\cdot Q[i]-Q[n]\}_{i=1}^n$}
    \State $i^*$ $\gets$ the first $i$ s.t. {\small $D[i-1]\leq 0\leq D[i]$}
    \State $s^*$ $\gets$ {\small $X[i^*]/\hat X[i^*]$}
    \State $l^*$ $\gets$ objective function value at $s^*$.
    \State \Return $s^*$, $l^*$.
\EndFunction
\end{algorithmic}
\end{algorithm}
\vspace{-20pt}
\end{figure}

\paravspace
\paragraph{Subproblem (\textit{w/} truncation).}
We truncate each residual term to suppress outliers. The truncated objective is
\begin{equation}
    \min_s l_1(s) = \min_s \sum_{i=1}^N\min(\tau, w_i|s\hat x_i-x_i|),
\label{eq:subproblem_w_truncate}
\end{equation}
where $\tau$ is set to $1$ in all our experiments. For each item $l_{1,i}(s)=\min(\tau, w_i|s\hat x_i-x_i|)$ in the equation, the one-sided derivatives are
\begin{equation}
\begin{split}
    l_{1,i}'^-(s)=\left\{\begin{array}{cc}
    -w_i\hat x_i&{w_ix_i-\tau_i\over w_i\hat x_i}<s\leq{x_i\over\hat  x_i}\\
    w_i\hat x_i&{x_i\over\hat  x_i}<s\leq{w_ix_i+\tau_i\over w_i\hat x_i}\\
    0&\text{otherwise}
    \end{array}\right.,
    \\
    l_{1,i}'^+(s)=\left\{\begin{array}{cc}
    -w_i\hat x_i&{w_ix_i-\tau_i\over w_i\hat x_i}\leq s<{x_i\over\hat  x_i}\\
    w_i\hat x_i&{x_i\over\hat  x_i}\leq s<{w_ix_i+\tau_i\over w_i\hat x_i}\\
    0&\text{otherwise}
    \end{array}\right..
\end{split}
\end{equation}
Therefore, the one-sided derivatives of $l_0(s)$ are
\begin{equation}
\begin{split}
    l_1'^{-}(s)
    &=\sum_{i=1}^N l_{1,i}'^-(s)
    =\!\!\!\!\sum_{{ x_i\over \hat x_i}<s\leq{w_i x_i+\tau\over w_i\hat x_i}}\!\!\!\!{w_i\hat x_i}-\!\!\!\!\sum_{{w_i x_i-\tau\over w_i\hat x_i}<s\leq{ x_i\over \hat x_i}}\!\!\!\!{w_i\hat x_i}
    \\
    &=2\sum_{{ x_i\over \hat x_i}<s}w_i\hat x_i-\!\!\!\!\sum_{{w_i x_i-\tau\over w_i\hat x_i}<s}\!\!\!\!w_i\hat x_i-\!\!\!\!\sum_{{w_i x_i+\tau\over w_i\hat x_i}<s}\!\!\!\!w_i\hat x_i,
\end{split}
\end{equation}
\vspace{-10pt}
\begin{equation}
\begin{split}
    l_1'^{+}(s)
    &=\sum_{i=1}^N l_{1,i}'^+(s)
    =\!\!\!\!\sum_{{ x_i\over \hat x_i}\leq s<{w_i x_i+\tau\over w_i\hat x_i}}\!\!\!\!{w_i\hat x_i}-\!\!\!\!\sum_{{w_i x_i-\tau\over w_i\hat x_i}\leq s<{ x_i\over \hat x_i}}\!\!\!\!{w_i\hat x_i}
    \\
    &=2\sum_{{ x_i\over \hat x_i}\leq s}w_i\hat x_i-\!\!\!\!\sum_{{w_i x_i-\tau\over w_i\hat x_i}\leq s}\!\!\!\!w_i\hat x_i-\!\!\!\!\sum_{{w_i x_i+\tau\over w_i\hat x_i}\leq s}\!\!\!\!w_i\hat x_i.
\end{split}
\end{equation} 

Lemma~\ref{lem:1} shows that the minimum of the objective function in Eq.~\ref{eq:subproblem_w_truncate} is still achieved at one of the points in the set $\{{x_i\over\hat  x_i}\}_{i=1}^N$, despite the function is non-convex and may contain local minima. 

Solving the subproblem requires two steps, as outlined in Algorithm~\ref{alg:subproblem_w_truncation}. 
The first step is to identify all extrema in $\{{x_i\over \hat 
x_i}\}_{i=1}^N$ that satisfy $l_0'^{-}({x_i\over\hat  x_i})< 0\leq l_0'^{+}({x_i\over\hat  x_i})$ by evaluating the derivative values. 
This can be done efficiently through first binary searching on the sorted arrays $\{{x_i\over\hat  x_i}\}_{i=1}^N$, $\{{w_ix_i-\tau\over w_i\hat x_i}\}_{i=1}^N$ and $\{{w_i x_i+\tau\over w_i\hat x_i}\}_{i=1}^N$ and then indexing the prefix summations of $\{w_i\hat x_i\}$ in the associated orders. 
This step has a complexity of $O(N\log N)$. 
The second step involves computing the objective values at these extrema and determining the minimum, which takes $O(Nn_e)$ time, where $n_e$ is the number of extrema. As $n_e$ approximates the number of outliers, it is typically a small constant in practice.

\begin{lemma}  
    \label{lem:1}  
    There exists at least one pair of $(k^*,s^*)$ such that $s^*\hat x_{k^*}-x_{k^*}=0$ and $s^*$ minimizes  Eq.~\ref{eq:subproblem_w_truncate}.
\end{lemma}
\paravspace
\begin{proof}
    The minimum of $l_1(s)$ must exist, because $l_1(s)$ is continuous, piece-wisely linear and bounded in $[0,N\tau]$. 
    
    We first prove that there must exist $s^*$ such that $l_1(s^*) = \min l_1(s)$ and $l_1'^+(s)>l_1'^-(s)$. Otherwise, for all $s^*$ such that $l_0(s^*)=\min l_1(s)$, there will be $l_1'^+(s)=l_1^-(s)=0$, hence the value of $l_1(s)$ in the linear interval where the minimum locates is constant. As a consequence, all neighboring intervals will be constant until the boundary where $\min l_1(s)=l_1(-\infty)=N\tau$, which contradicts the obvious fact that $\min l_1(s) \leq l_1(x_1/\hat x_1) < N\tau$. 
    
    Given $l_0'^+(s^*)>l_0'^-(s^*)$, there exists an index $k^*$ such that $s^*=\hat x_{k^*}/x_{k^*}$, because
    \begin{equation}
    \begin{split}
        0<&l_1'^+(s^*)-l_1'^-(s^*)
        \\
        =&2\sum_{{x_i\over\hat  x_i}=s^*}w_i\hat x_i-\!\!\!\!\!\!\sum_{{w_ix_i-\tau\over w_i\hat x_i}=s^*}\!\!\!\!w_i\hat x_i-\!\!\!\!\!\!\sum_{{w_i x_i+\tau\over w_i\hat x_i}=s^*}\!\!\!\!w_i\hat x_i
        \\
        \leq &2\sum_{{x_i\over\hat  x_i}=s^*}w_i\hat x_i.
    \end{split}
    \end{equation}
\end{proof}

\begin{figure}[H]
\vspace{-25pt}
\begin{algorithm}[H]
\caption{ROE alignment subproblem \textit{w/} truncation}
\label{alg:subproblem_w_truncation}
\begin{algorithmic}    
\State \textbf{input:} arrays $\hat X[1..n]$, $X[1..n]$, $W[1..n]$, float $\tau$
\State \textbf{output:} the optimal scale $s^*$, objective value $l^*$ to Eq.~\ref{eq:subproblem_w_truncate}
\Statex
\vspace{-5pt}
\Function{SolveSubproblem}{\small $\hat X$, $X$, $W$, $\tau$}
	\State $A[1..n]$ $\gets$ {\small $X/\hat X$}
    \State $B[1..n]$ $\gets$ {\small $(W*X-\tau)/(W*\hat X)$}
    \State $C[1..n]$ $\gets$ {\small $(W*X+\tau)/(W*\hat X)$}
    \For {each array $\mathcal A$ in $\{A,B,C\}$}
        \State sort $\mathcal A$ and obtain sorted indices $I_\mathcal A[1..n]$
        \State $Q_\mathcal A[1..n]$ $\gets$ accumulated sum of {\small $\{W\hat X[I_\mathcal A[i]]\}_{i=1}^n$}
    \EndFor
    
    \State Initialize $I_E$ as empty set
    \For {$i=1$ to $n$} \Comment{\emph{parallel computation}}
        \For {each array $\mathcal A$ in $\{A,B,C\}$}
            \State $j_\mathcal A^-$ $\gets$ the last $j$ s.t. $\mathcal A[j] < X[i]/\hat X[i]$
            \State $j_\mathcal A^+$ $\gets$ the last $j$ s.t. $\mathcal A[j] \leq X[i]/\hat X[i]$
        \EndFor
        \State $d^-$ $\gets$ $2\cdot Q_A[j_A^-] - Q_B[j_B^-]-Q_C[j_C^-]$
        \State $d^+$ $\gets$ $2\cdot Q_A[j_A^+] - Q_B[j_B^+]-Q_C[j_C^+]$
        \State \textbf{if} {$d^-<0\leq d^+$} \textbf{then} append $i$ to $I_E$
    \EndFor

    \State Initialize $l[1..N]$ with $\infty$
    \For {$i$ in $I_E$} \Comment{\emph{parallel computation}}
        \State $s$ $\gets$ $X[i]/\hat X[i]$
        \State $l[i]$ $\gets$ objective function value at $s$
    \EndFor
    \State $i^*$ $\gets$ index of the minimum in $l[i],i\in I_E$
    \State $s^*$ $\gets$ $X[i^*]/\hat X[i^*]$, $l^*$ $\gets$ $l[i^*]$
    \State \Return $s^*$, $l^*$
\EndFunction
\end{algorithmic}
\end{algorithm}
\vspace{-20pt}
\end{figure}

\paravspace
\paragraph{Alignment with 1D shift.}

Recall the alignment objective and let $w_i$ be $1/z_i$. 
We rewrite it as follows:
\begin{equation}
\begin{split}
    \min_{s,t_z} \sum_{i=1}^n  [w_i|s\hat x_i\!-x_i| +w_i|s\hat y_i\!-\!y_i|+ w_i|s\hat z_i+t_z-z_i|],
    \label{eq:global_align_rewrite_wo_truncation}
\end{split}
\end{equation}
or apply truncation to each absolute residual term
\begin{equation}
\begin{split}
    \min_{s,t_z} \sum_{i=1}^n [\min(\tau, w_i|s\hat x_i\!-x_i|)\! +\!\min(\tau, w_i|s\hat y_i\!-\!y_i|) \\+ \min(\tau, w_i|s\hat z_i+t_z-z_i|)].
    \label{eq:global_align_rewrite_w_truncation}
\end{split}
\end{equation}
The proposed solution is outlined in Algorithm~\ref{alg:solver_1d}, with proof as follows. The corresponding subproblem solver is selected  based on whether truncation is applied.

\begin{lemma}  
    \label{lem:2}  
    There exists at least one triplet of $(k^*,s^*,t_z^*)$ such that $s^*\hat z_{k^*}+t_z^*-z_{k^*}=0$ and $(s^*,t_z^*)$ minimizes the objective of Equation~\ref{eq:global_align_rewrite_w_truncation}. 
\end{lemma}
\paravspace
\begin{proof}  
    Denote the objective as $l_2(s,t_z)$, 
    \begin{equation}
    \begin{split}
        l_2(s,t_z)&=\sum_{i=1}^n[\min\left(\tau, w_i{|s\hat x_i-x_i|}\right) + \min\left(\tau, w_i{|s\hat y_i-y_i|}\right)] \\&+ \sum_{i=1}^N \min\left(\tau, w_i{|t_z-(z_i-s\hat z_i)|}\right)].
    \end{split}
    \end{equation}
    Given arbitrary $s$, using Lemma~\ref{lem:1}, there exists at least one pair $(t_z,k)$ such that $t_z-(z_k-s\hat z_k)=0$ and $t_z$ minimizes $\sum_{i=1}^n\min\left(\tau, w_i{|s\hat z_i+t_z-z_i|}\right)$, hence minimizes $l_2(s,t_z)$ as the rest parts are constant with regard to $t_z$. Therefore, a solution $s^*$ is always associated with corresponding $(t^*_z,k^*)$ such that that $s^*\hat z_{k^*}+t^*_z-z_{k^*}=0$. 
\end{proof}

Lemma~\ref{lem:2} allows us to reduce Eq.~\ref{eq:global_align_rewrite_w_truncation} to the subproblem with respect to some index $k$. For each possible index $k$, the objective is formed as:
\begin{equation}
\begin{split}
    \min_s \sum_{i=1}^n \min(\tau, w_i|s\hat x_i\!-\!sx_i|)\! +\!\min(\tau, w_i|s\hat y_i\!-\!y_i|) \\+ \min(\tau, w_i|s(\hat z_i-\hat z_k)-(z_i-z_k)|),
\end{split}
\end{equation}
which is solvable in $O(N\log N)$ complexity. We enumerate all possible indices for $k$ and find the minimum. Therefore, the total time complexity is $O(N^2\log N)$. 

In our implementation, the point map is resized to low resolution ($64\times64$) for alignment, with $N=4096$ at most. The algorithm is further parallelized with tensor operations on GPUs.

\begin{algorithm}[h]
\caption{ROE alignment \textit{w/} 1-D shift}
\label{alg:solver_1d}
\begin{algorithmic}    
\State \textbf{input:} point arrays $\hat P[1..N,1..3]$, $P[1..N,1..3]$, 
\State \quad\quad\quad weight array $W[1..N]$ 
\State \textbf{output:} the optimal scale $s^*$, shift $t_z^*$, 
\State \quad\quad\quad objective value $l^*$ to  Eq.~\ref{eq:global_align_rewrite_wo_truncation} or Eq.~\ref{eq:global_align_rewrite_w_truncation} 
\Statex
\vspace{-5pt}
\State {$W[1..3N]$ $\gets$ repeat each element in $W$ 3 times}
\State Initialize arrays $s[1..N]$, $l[1..N]$, $t_z[1..N]$
\For {$k = 1$ to $N$} \Comment{\emph{parallel computation}} 
    \State {\small $\hat X[1...3N]$$\gets$ \Call{Flatten}{$\hat P[1..N,1..3]-\{0,0,\hat P[k,3]\}$}} 
    \State {\small $X[1...3N]$$\gets$ \Call{Flatten}{$P[1..N,1..3]-\{0,0,P[k,3]\}$}} 
    \State $(s[k], l[k])$ $\gets$ \Call{SolveSubproblem}{$\hat X$,$X$,$W$}
    \State $t_z[k]$ $\gets$ $P[k, 3] - s[k]\cdot\hat P[k, 3]$
\EndFor

\State $k^*$ $\gets$ index of the minimum in $l[1..N]$
\State $s^*$ $\gets$ $s[k^*]$, $l^*$ $\gets$ $l[k^*]$, $t_z^*$ $\gets$ $t_z[k^*]$
\State \Return $s^*$, $t_z^*$, $l^*$
\end{algorithmic}
\end{algorithm}

\paravspace
\paragraph{Alignment with 3D shift.}

We apply truncation and rewrite the objective as follows:

\begin{equation}
\begin{split}
    \min_{s,t_z} \sum_{i=1}^n &\min(\tau, w_i|s\hat x_i\!+t_x-x_i|)\\&+\min(\tau, w_i|s\hat y_i+t_y-y_i|) \\&+ \min(\tau, w_i|s\hat z_i+t_z-z_i|).
    \label{eq:align3d_rewrite_w_truncation}
\end{split}
\end{equation}

Similarly to the proof of Lemma~\ref{lem:2}, there exists at least one group $(k_1^*, k_2^*, k_3^*, s^*, \mathbf{t}^*)$ such that $s^*\hat x_{k_1^*} + t_x^* - x_{k_1^*} = 0$, $s^*\hat y_{k_2^*} + t_y^* - y_{k_2^*} = 0$, $s^*\hat z_{k_3^*} + t_z^* - z_{k_3^*} = 0$, and $(s^*, \mathbf{t}^*)$ minimizes the objective. However, the $O(N^4 \log N)$ time complexity of a brute-force search is prohibitive. Motivated by the strong locality of surface points within a 3D sphere, we introduce a reasonable assumption, $k_1 = k_2 = k_3$, to obtain an approximately optimal solution with $O(N^2 \log N)$ complexity. This assumption posits that the predicted and ground truth patches can be well aligned under the condition that one corresponding pair of points coincides. The effectiveness of the approximated solution has been empirically validated.

\begin{algorithm}[H]
\caption{ROE alignment \textit{w/} 3-D shift}
\label{alg:solver_3d}
\begin{algorithmic}    
\State \textbf{input:} point arrays $\hat P[1..N,1..3]$, $P[1..N,1..3]$, 
\State \quad\quad\quad weight array $W[1..N]$ 
\State \textbf{output:} the optimal scale $s^*$, shift $\mathbf t^*$,
\State \quad\quad\quad objective value $l^*$ to  Eq.~\ref{eq:align3d_rewrite_w_truncation}
\Statex
\vspace{-5pt}
\State {$W[1..3N]$ $\gets$ repeat each element in $W$ 3 times}
\State Initialize arrays $s[1..N]$, $l[1..N]$, $\mathbf t[1..N, 3]$
\For {$k = 1$ to $N$} \Comment{\emph{parallel computation}} 
    \State {$\hat X[1...3N]$$\gets$ \Call{Flatten}{$\hat P[1..N,1..3]-\hat P[k,1..3]$}}
    \State {$X[1...3N]$$\gets$ \Call{Flatten}{$P[1..N,1..3]-P[k,1..3]$}} 
    \State $(s[k], l[k])$ $\gets$ \Call{SolveSubproblem}{$\hat X$,$X$,$W$}
    \State $\mathbf t[k]$ $\gets$ $P[k] - s[k]\cdot\hat P[k]$
\EndFor

\State $k^*$ $\gets$ index of the minimum in $l[1..N]$
\State $s^*$ $\gets$ $s[k^*]$, $l^*$ $\gets$ $l[k^*]$, $\mathbf t^*$ $\gets$ $\mathbf t[k^*]$
\State \Return $s^*$, $\mathbf t^*$, $l^*$
\end{algorithmic}
\end{algorithm}

\section{Experiment Details}

\subsection{Training Data}

The datasets used in our training are listed in Table~\ref{tab:datasets}. The number of frames may slightly differ from that of the original data because some invalid frames are dropped. 

To assign balanced weights to the datasets for training, we compute the retrieval probability of each dataset relative to OpenImagesV7~\cite{OpenImages}, a large and diverse natural image dataset. Specifically, we leverage DINOv2~\cite{oquab2023dinov2} to extract feature vectors and calculate the probability that the nearest neighbor of a randomly selected image from OpenImagesV7 is found in each respective training dataset.

\begin{table}[h]
    \centering
    \setlength{\tabcolsep}{5pt}
    \footnotesize
    \begin{tabular}{l c c c c}
        \specialrule{0.12em}{0em}{0em}
        Name & Domain & \#Frames & Type & Weight \\
        \hline
        A2D2\cite{geyer2020a2d2} & Outdoor/Driving & $196$K & C & 0.8\%\\
        
        Argoverse2\cite{Argoverse2} & Outdoor/Driving & $1.1$M & C & 7.4\%\\
        
        ARKitScenes\cite{dehghan2021arkitscenes} & Indoor & $449$K & B & 8.6\%\\
        
        DIML-indoor\cite{diml} & Indoor & $894$K & D & 4.8\%\\
        
        BlendedMVS\cite{yao2020blendedmvs} & In-the-wild & $115$K & B & 12.0\%\\
        
        MegaDepth\cite{MegaDepthLi18} & Outdoor/In-the-wild & $92$K & B & 5.6\%\\
        
        Taskonomy\cite{zamir2018taskonomy} & Indoor & $3.6$M & B & 14.1\%\\
        
        Waymo\cite{sun2020waymo} & Outdoor/Driving & $788$K & C & 6.4\%\\
        
        GTA-SfM\cite{Wang2019gtasfm} & Outdoor/In-the-wild & $19$K & A & 2.8\% \\
        
        Hypersim\cite{roberts2021hypersim} & Indoor & $75$K & A & 5.0\%\\
        
        IRS\cite{wang2019IRS} & Indoor & $101$K & A & 5.6\%\\
        
        KenBurns\cite{niklaus2019kenburns} & In-the-wild & $76$K & A & 1.6\%\\
        
        MatrixCity\cite{li2023matrixcity} & Outdoor/Driving & $390$K & A &1.3\%\\
        
        MidAir\cite{Fonder2019MidAir} & Outdoor/In-the-wild & $423$K & A &4.0\%\\
        
        MVS-Synth\cite{huang2018mvsynth} & Outdoor/Driving & $12$K & A &1.2\%\\
        
        Spring\cite{Mehl2023Spring} & In-the-wild & $5$K & A &0.7\%\\
        
        Structured3D\cite{Structured3D} & Indoor & $77$K & A &4.8\%\\
        
        Synthia\cite{Ros2016synthia} & Outdoor/Driving & $96$K & A &1.2\%\\
        
        TartanAir\cite{tartanair2020iros} & In-the-wild & $306$K & A &5.0\%\\
        
        UrbanSyn\cite{gómez2023urbansyn} & Outdoor/Driving & $7$K & A &2.1\%\\
        
        ObjaverseV1\cite{objaverse} & Object & $167$K & A & 4.8\%\\
        \specialrule{0.12em}{0em}{0em}
        \vspace{5pt}
    \end{tabular}

    \centering
    \scriptsize
    \setlength{\tabcolsep}{0.8pt}
    \begin{tabular}{l|ccc|cccccc}
        \specialrule{0.12em}{0em}{0em}
        
        \multirow{2}{*}{\footnotesize Type} & \multicolumn{3}{c|}{\footnotesize  Label quality}& \multicolumn{6}{c}{\footnotesize  Applied losses}\\
        
        &Accuracy &Range &Density &$\mathcal L_G$ &$\mathcal L_{S_1}$ &$\mathcal L_{S_2}$ &$\mathcal L_{S_3}$ &$\mathcal L_N$ &$\mathcal L_M$\\ 
        
        \hline
        A. Synthetic & Perfect   & $\infty$  & Dense  & \ding{51} & \ding{51} & \ding{51} & \ding{51} & \ding{51} & \ding{51}\\
        
        B. SfM/MV Recon & High  & $\infty$  & Dense\&Partial & \ding{51} & \ding{51} & \ding{51}  &  & & \ding{51} \\
        
        C. LiDAR/Laser & High    & $\sim100$m & Sparse & \ding{51} & \ding{51} &   &  & & \ding{51}\\
        
        D. Kinect & Medium & $\sim10$m & Dense& \ding{51} &  &  &  &  & \ding{51}\\
        \specialrule{0.12em}{0em}{0em}
    \end{tabular}
    \caption{Datasets used for training and tailored loss combination.}
    \label{tab:datasets}
\end{table}

\subsection{Evaluation Data}

The raw evaluation datasets are processed accordingly for reliable evaluation and fair comparison. We report the details as follows.

\begin{figure}[t]
    \centering
    \includegraphics[width=\linewidth]
    {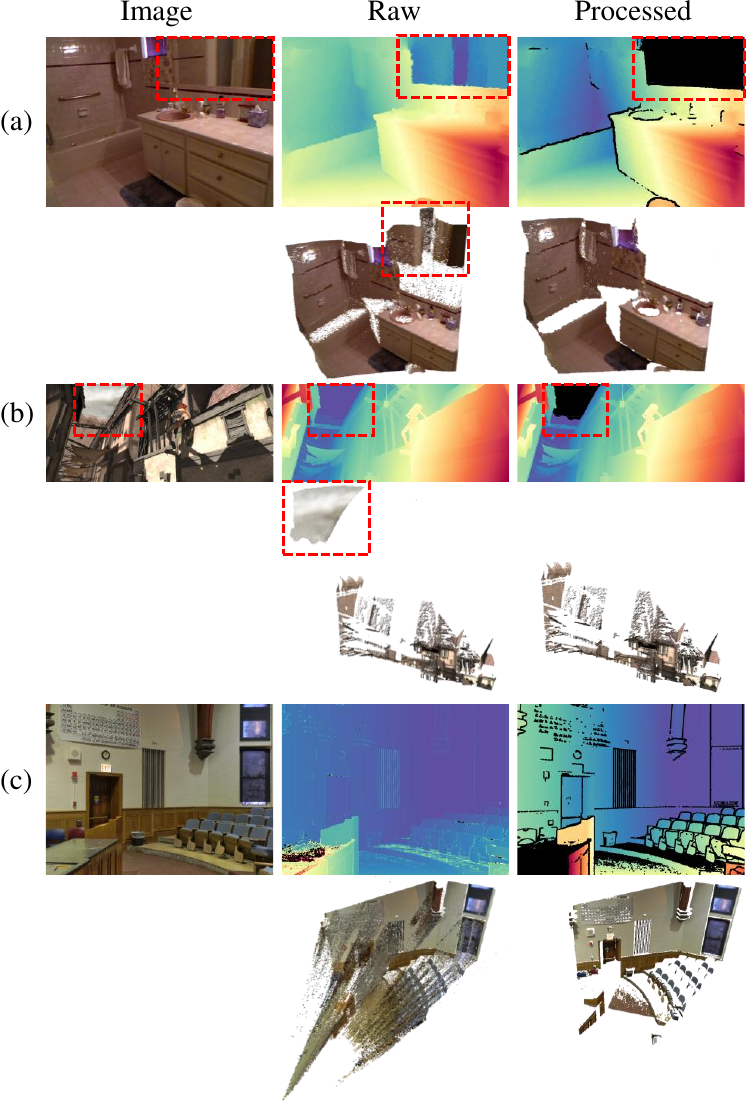}
    \vspace{-20pt}
    \caption{Examples of evaluation data preprocessing: (a) Removing mirror and boundary artifacts from the ground truth depth in NYUv2. (b) Excluding sky regions in Sintel. (c) Removing boundary artifacts from the ground truth depth in DIODE.}
    \label{fig:dataset_processing}
    \vspace{-20pt}
\end{figure}

\begin{itemize}
    \item \textbf{NYUv2~\cite{Silberman2012nyuv2}.} We use the official test split of 654 samples. Due to the inaccuracy of ground truth values captured by Kinect V1 near boundaries, we filter and remove boundary regions by a simple edge detection method. Specifically, we set a threshold for the difference between the minimum and maximum depth values within a local window. Depth values beyond 5 meters are excluded because they are unreliable due to the limited sensor range~\cite{tolgyessy2021kinect}. Additionally, we manually mask out areas with reflective and transparent surfaces, such as mirrors and glass, which cannot be accurately captured by the sensor.

    \item \textbf{KITTI~\cite{Uhrig2017kitti}.} We utilize the test split of 652 images of Eigen \etal~\cite{eigen2014depth} following previous works. The original resolution of $1242\times375$ does not match our training aspect ratio (ranging from $1:2$ to $2:1$), so we apply center cropping to obtain a resolution of $750\times375$ from the raw images.

    \item \textbf{ETH3D~\cite{Schops2019ETH3D}.} All 454 images are included. The images are undistorted with the official calibration data and downsized from the original resolution of $6202\times4135$ to $2048\times1365$.

    \item \textbf{iBims-1~\cite{ibim1_1}.} All 100 images are included at an original resolution of $640\times480$. 

    \item \textbf{GSO~\cite{downs2022googlescannedobjects}.} The dataset contains 1,030 objects. For each object, we render a single view at $512\times512$ resolution. The view is randomly sampled with a FOV ranging from $30^\circ$ to $60^\circ$. The object is centered in the image, and its bounding box occupies approximately 70\% of the image's size.

    \item \textbf{Sintel~\cite{Butler2012sintel}.} We use all 1,064 frames and center-crop the images to $872\times436$ from the original $1024\times436$ resolution to fit our aspect ratio range. The sky regions are manually masked out because evaluating models with sky depth included is not meaningful.

    \item \textbf{DDAD~\cite{ddadpacking}.} We randomly select 1,000 samples from the validation set. The dataset was collected using multiple cameras and LiDAR sensors mounted on a moving vehicle. Some cameras inadvertently capture parts of the vehicle, causing discrepancies with the sensor's depth data. To address this issue, we crop the regions that are not obstructed by the vehicle itself.

    \item \textbf{DIODE~\cite{diode_dataset}.} We utilize the official validation split, which includes 325 indoor images and 446 outdoor images at an original resolution of $1024\times768$. Due to artifacts in ground truth depth values near the boundaries in this dataset, we identify and remove these boundary regions using a similar approach as described above.
\end{itemize}

\subsection{Evaluation Protocol}

For all our models and baselines, predictions and ground truth are aligned in scale (and shift, if applicable) for each image before measuring errors. To clarify the notations in this section:
\begin{itemize}
    \item $\hat{\mathbf{p}}_i$ and $\mathbf{p}_i$ are the predicted and ground-truth points, respectively.
    \item $\hat z_i$ and $z_i$ are the predicted and ground-truth depths, which are the $Z$-coordinate of corresponding points.
    \item $\mathcal M$ is the mask of valid ground-truth.
    \item $a$ and $b$ denote the scale and shift used to align predictions with the ground truth for evaluation, to avoid confusion with similar symbols used in the training objectives.
\end{itemize}

\begin{itemize}
    \item \textbf{Scale-invariant point map.} The scale $a^*$ to align prediction with ground truth is computed as:
    \begin{equation}
        a^* = \mathop{\arg\!\min}_a \sum_{i\in\mathcal M} {1\over z_i}\|a\hat {\mathbf p}_i-\mathbf p_i\|_1,
    \end{equation}

    \item \textbf{Affine-invariant point map.} The scale $a^*$ and shift $\mathbf b^*$ are computed as:
    \begin{equation}
        (a^*,\mathbf b^*) = \mathop{\arg\!\min}_{a,\mathbf b} \sum_{i\in\mathcal M} {1\over z_i}\|a\hat {\mathbf p}_i+\mathbf b-\mathbf p_i\|_1.
    \end{equation}
    
    \item \textbf{Scale-invariant depth map}, the scale $a^*$ is computed as 
    \begin{equation}
        a^* = \mathop{\arg\!\min}_s \sum_{i\in\mathcal M} {1\over z_i}|a\hat z_i-z_i|.
    \end{equation}

    \item \textbf{Affine-invariant depth map.} The scale $a^*$ and shift $b^*$ are computed as 
    \begin{equation}
        (a^*, b^*) = \mathop{\arg\!\min}_s \sum_{i\in\mathcal M} {1\over z_i}|a\hat z_i+b-z_i|.
    \end{equation}

    \item \textbf{Affine-invariant disparity map.} We follow the established protocol for affine disparity alignment~\cite{ranftl2020midas}, using least-squares to align predictions in disparity space:
    \begin{equation}
        (a^*, b^*) = \mathop{\arg\!\min}_s \sum_{i\in\mathcal M}(a\hat d_i+b-d_i)^2,
    \end{equation}
    where $\hat d_i$ is the predicted disparity and $d_i$ is the ground truth, defined as $d_i=1/z_i$. 
    To prevent aligned disparities from taking excessively small or negative values, the aligned disparity is truncated by the inverted maximum depth $1/z_\text{max}$ before inversion. The final aligned depth $\hat z_i^*$ is computed as:
    \begin{equation}
        \hat z_i^* := {1\over \max(a^* \hat d_i + b^*, 1/z_\text{max})}.
    \end{equation}

\end{itemize}

\section{More Results}

\paragraph{Full table of depth estimation results}

In Table~\ref{tab:comparison_depth_full}, we present detailed results for depth estimation where methods that predict metric or scale-invariant depth are also evaluated on affine-invariant depth and disparity for a fair comparison.

\begin{table*}[t]
    \centering
    \setlength{\tabcolsep}{2.5pt}
    \begin{tabular}{l|cc|cc|cc|cc|cc|cc|cc|cc|ccc}
        \specialrule{.12em}{0em}{0em}
        \multirow{2}{*}{\textbf{Method}} & \multicolumn{2}{c|}{NYUv2} & \multicolumn{2}{c|}{KITTI} & \multicolumn{2}{c|}{ETH3D} & \multicolumn{2}{c|}{iBims-1} & \multicolumn{2}{c|}{GSO} & \multicolumn{2}{c|}{Sintel} & \multicolumn{2}{c|}{DDAD} & \multicolumn{2}{c|}{DIODE} & \multicolumn{3}{c}{Average}\\
        & \scriptsize Rel\textsuperscript{d}\scriptsize$\downarrow$ & \scriptsize $\delta_1^\text{d}$\scriptsize$\uparrow$ & \scriptsize Rel.\scriptsize$\downarrow$ & \scriptsize $\delta_1^\text{d}$\scriptsize$\uparrow$ & \scriptsize Rel\textsuperscript{d}\scriptsize$\downarrow$ & \scriptsize $\delta_1^\text{d}$\scriptsize$\uparrow$ & \scriptsize Rel.\scriptsize$\downarrow$ & \scriptsize $\delta_1^\text{d}$\scriptsize$\uparrow$ & \scriptsize Rel\textsuperscript{d}\scriptsize$\downarrow$ & \scriptsize $\delta_1^\text{d}$\scriptsize$\uparrow$ & \scriptsize Rel\textsuperscript{d}\scriptsize$\downarrow$ & \scriptsize $\delta_1^\text{d}$\scriptsize$\uparrow$ & \scriptsize Rel\textsuperscript{d}\scriptsize$\downarrow$ & \scriptsize $\delta_1^\text{d}$\scriptsize$\uparrow$ & \scriptsize Rel\textsuperscript{d}\scriptsize$\downarrow$ & \scriptsize $\delta_1^\text{d}$\scriptsize$\uparrow$ & \scriptsize Rel\textsuperscript{d}\scriptsize$\downarrow$ & \scriptsize $\delta_1^\text{d}$\scriptsize$\uparrow$ & \scriptsize Rank\scriptsize$\downarrow$\\
        \hline
        \multicolumn{19}{c}{Scale-invariant depth}\\
        \hline
        
LeReS & 12.1 & 82.6 & 19.2 & 64.8 & 14.2 & 78.4 & 14.0 & 78.8 & 13.6 & 77.9 & 30.5 & 52.1 & 26.5 & 52.0 & 18.2 & 69.6 & 18.5 & 69.5 & 7.31 \\

ZoeDepth & \textcolor{lightgray}{5.62} & \textcolor{lightgray}{96.3} & \textcolor{lightgray}{7.27} & \textcolor{lightgray}{91.9} & 10.4 & 87.3 & 7.45 & 93.2 & 3.23 & \underline{99.9} & 27.4 & 61.8 & 17.0 & 72.8 & 11.3 & 85.2 & \textcolor{lightgray}{11.2} & \textcolor{lightgray}{86.1} & 5.50 \\

DUSt3R & 4.40 & 97.1 & 7.81 & 90.6 & 6.04 & 95.7 & 4.98 & 95.8 & 3.27 & 99.5 & 31.1 & 57.2 & 18.6 & 73.3 & 8.91 & 88.8 & 10.6 & 87.2 & 5.00 \\

Metric3D V2 & 4.69 & 97.4 & \underline{4.00} & \underline{98.5} & \underline{3.84} & \underline{98.5} & \underline{4.23} & \underline{97.7} & \underline{2.46} & \underline{99.9} & \underline{20.7} & \underline{69.8} & \textcolor{lightgray}{7.41} & \textcolor{lightgray}{94.6} & \textbf{3.29} & \textbf{98.4} & \textcolor{lightgray}{6.33} & \textcolor{lightgray}{94.3} & \underline{2.07} \\

UniDepth & \underline{3.86} & \textbf{98.4} & \textbf{3.73} & \textbf{98.6} & 5.67 & 97.0 & 4.79 & 97.4 & 4.18 & 99.7 & 28.3 & 58.8 & \underline{10.1} & \textbf{90.5} & 6.83 & 92.8 & \underline{8.43} & \underline{91.6} & {3.00} \\

DA V1 & \textcolor{lightgray}{4.77} & \textcolor{lightgray}{97.5} & \textcolor{lightgray}{5.61} & \textcolor{lightgray}{95.6} & 9.41 & 88.9 & 5.53 & 95.8 & 5.49 & 99.3 & 28.3 & 56.7 & 13.2 & 81.5 & 10.3 & 87.5 & \textcolor{lightgray}{10.3} & \textcolor{lightgray}{87.9} & 5.67 \\

\quad\scriptsize{-metric indoor} & \scriptsize{\textcolor{lightgray}{4.77}} & \scriptsize{\textcolor{lightgray}{97.5}} & \scriptsize{15.4} & \scriptsize{73.6} & \scriptsize{9.41} & \scriptsize{88.9} & \scriptsize{5.53} & \scriptsize{95.8} & \scriptsize{5.49} & \scriptsize{99.3} & \scriptsize{28.3} & \scriptsize{56.7} & \scriptsize{24.2} & \scriptsize{57.4} & \scriptsize{10.3} & \scriptsize{87.5} & \scriptsize{\textcolor{lightgray}{12.9}} & \scriptsize{\textcolor{lightgray}{82.1}} & - \\

\quad\scriptsize{-metric outdoor} & \scriptsize{15.9} & \scriptsize{72.3} & \scriptsize{\textcolor{lightgray}{5.61}} & \scriptsize{\textcolor{lightgray}{95.6}} & \scriptsize{8.77} & \scriptsize{92.4} & \scriptsize{13.8} & \scriptsize{78.8} & \scriptsize{8.59} & \scriptsize{93.6} & \scriptsize{28.1} & \scriptsize{54.8} & \scriptsize{13.2} & \scriptsize{81.5} & \scriptsize{13.0} & \scriptsize{81.4} & \scriptsize{\textcolor{lightgray}{13.4}} & \scriptsize{\textcolor{lightgray}{81.3}} & -\\

DA V2 & 5.03 & 97.3 & 7.23 & 93.7 & 6.12 & 95.5 & 4.32 & \textbf{97.9} & 4.38 & 99.3 & 23.0 & 65.2 & 14.7 & 78.0 & 7.95 & 90.0 & 9.09 & 89.6 & 4.06 \\

\quad\scriptsize{-metric indoor} & \scriptsize{5.03} & \scriptsize{97.3} & \scriptsize{7.61} & \scriptsize{90.9} & \scriptsize{6.12} & \scriptsize{95.5} & \scriptsize{4.32} & \scriptsize{97.9} & \scriptsize{4.38} & \scriptsize{99.3} & \scriptsize{23.0} & \scriptsize{65.2} & \scriptsize{16.6} & \scriptsize{73.4} & \scriptsize{7.95} & \scriptsize{90.0} & \scriptsize{9.38} & \scriptsize{88.7} & - \\

\quad\scriptsize{-metric outdoor} & \scriptsize{15.3} & \scriptsize{72.3} & \scriptsize{7.23} & \scriptsize{93.7} & \scriptsize{9.30} & \scriptsize{89.6} & \scriptsize{10.6} & \scriptsize{84.9} & \scriptsize{9.62} & \scriptsize{92.5} & \scriptsize{28.6} & \scriptsize{57.3} & \scriptsize{14.7} & \scriptsize{78.0} & \scriptsize{12.2} & \scriptsize{83.2} & \scriptsize{13.4} & \scriptsize{81.4} & - \\

\textbf{Ours} & \textbf{3.44} & \textbf{98.4} & 4.25 & 97.8 & \textbf{3.36} & \textbf{98.9} & \textbf{3.46} & 97.0 & \textbf{1.47} & \textbf{100} & \textbf{19.3} & \textbf{73.4} & \textbf{9.17} & \textbf{90.5} & \underline{4.89} & \underline{94.7} & \textbf{6.17} & \textbf{93.8} & \textbf{1.62} \\
        
        \hline
        \multicolumn{19}{c}{Affine-invariant depth}\\
        \hline
        
LeReS & 6.21 & 95.4 & 8.28 & 90.3 & 8.95 & 90.8 & 6.68 & 94.5 & 4.03 & 99.4 & 24.0 & 64.8 & 16.2 & 75.8 & 9.99 & 88.1 & 10.5 & 87.4 & 8.81 \\

ZoeDepth & \textcolor{lightgray}{4.76} & \textcolor{lightgray}{97.3} & \textcolor{lightgray}{5.59} & \textcolor{lightgray}{95.1} & 7.27 & 94.2 & 5.85 & 95.7 & 2.54 & 99.9 & 21.8 & 69.2 & 14.2 & 80.1 & 7.80 & 90.9 & \textcolor{lightgray}{8.73} & \textcolor{lightgray}{90.3} & 7.33 \\

DUSt3R & 3.73 & 97.8 & 7.30 & 91.6 & 4.96 & 96.4 & 3.94 & 96.6 & 2.55 & 99.6 & 25.4 & 64.2 & 16.9 & 76.2 & 6.68 & 92.6 & 8.93 & 89.4 & 6.62 \\

Metric3D V2 & 3.94 & 97.6 & \textbf{3.50} & \underline{98.4} & \underline{3.24} & \underline{99.0} & \underline{3.28} & \textbf{98.3} & 2.10 & 99.4 & 26.6 & 71.7 & \textcolor{lightgray}{7.15} & \textcolor{lightgray}{94.8} & \textbf{2.75} & \textbf{98.7} & \textcolor{lightgray}{6.57} & \textcolor{lightgray}{94.7} & 3.64 \\

UniDepth V1 & \underline{3.40} & \textbf{98.6} & \underline{3.55} & \textbf{98.7} & 4.92 & 97.5 & 3.76 & 98.2 & 2.48 & 99.9 & 24.9 & 64.1 & \underline{9.46} & \underline{90.8} & 4.90 & 96.2 & 7.17 & \underline{93.0} & 3.62 \\

Marigold & 4.63 & 97.3 & 7.29 & 93.8 & 6.08 & 96.3 & 4.35 & 97.2 & 2.78 & 99.9 & 21.2 & 75.0 & 14.6 & 80.5 & 6.34 & 94.3 & 8.41 & 91.8 & 5.69 \\

Geowizard & 4.69 & 97.4 & 8.14 & 92.5 & 6.90 & 93.9 & 4.50 & 97.1 & 2.00 & 99.9 & 17.8 & 76.2 & 16.5 & 75.7 & 7.03 & 92.7 & 8.45 & 90.7 & 6.44 \\

DA V1 & \textcolor{lightgray}{3.82} & \textcolor{lightgray}{98.3} & \textcolor{lightgray}{5.04} & \textcolor{lightgray}{96.4} & 6.23 & 95.2 & 4.23 & 97.3 & 1.98 & \textbf{100} & 20.1 & 71.8 & 11.3 & 86.1 & 6.75 & 92.6 & \textcolor{lightgray}{7.43} & \textcolor{lightgray}{92.2} & 4.83 \\

\quad\scriptsize{-metric indoor} & \scriptsize{\textcolor{lightgray}{3.82}} & \scriptsize{\textcolor{lightgray}{98.3}} & \scriptsize{9.95} & \scriptsize{86.5} & \scriptsize{6.23} & \scriptsize{95.2} & \scriptsize{4.23} & \scriptsize{97.3} & \scriptsize{1.98} & \scriptsize{100} & \scriptsize{20.1} & \scriptsize{71.8} & \scriptsize{17.0} & \scriptsize{74.0} & \scriptsize{6.75} & \scriptsize{92.6} & \scriptsize{\textcolor{lightgray}{8.76}} & \scriptsize{\textcolor{lightgray}{89.5}} & - \\

\quad\scriptsize{-metric outdoor} & \scriptsize{7.68} & \scriptsize{93.8} & \scriptsize{\textcolor{lightgray}{5.04}} & \scriptsize{\textcolor{lightgray}{96.4}} & \scriptsize{6.21} & \scriptsize{96.6} & \scriptsize{7.00} & \scriptsize{94.2} & \scriptsize{2.77} & \scriptsize{99.8} & \scriptsize{20.6} & \scriptsize{70.0} & \scriptsize{11.3} & \scriptsize{86.1} & \scriptsize{7.03} & \scriptsize{93.2} & \scriptsize{\textcolor{lightgray}{8.45}} & \scriptsize{\textcolor{lightgray}{91.3}} & - \\

DA V2 & 4.16 & 97.9 & 6.77 & 94.3 & 4.63 & 97.2 & 3.44 & \textbf{98.3} & \underline{1.44} & \textbf{100} & \underline{17.1} & \underline{76.6} & 13.4 & 81.8 & 5.41 & 94.6 & \underline{7.04} & 92.6 & - \\

\quad\scriptsize{-metric indoor} & \scriptsize{4.16} & \scriptsize{97.9} & \scriptsize{7.09} & \scriptsize{92.3} & \scriptsize{4.63} & \scriptsize{97.2} & \scriptsize{3.44} & \scriptsize{98.3} & \scriptsize{1.44} & \scriptsize{100} & \scriptsize{17.1} & \scriptsize{76.6} & \scriptsize{14.3} & \scriptsize{79.8} & \scriptsize{5.41} & \scriptsize{94.6} & \scriptsize{7.20} & \scriptsize{92.1} & - \\

\quad\scriptsize{-metric outdoor} & \scriptsize{8.65} & \scriptsize{91.0} & \scriptsize{6.77} & \scriptsize{94.3} & \scriptsize{7.24} & \scriptsize{93.5} & \scriptsize{6.80} & \scriptsize{93.5} & \scriptsize{2.29} & \scriptsize{100} & \scriptsize{22.4} & \scriptsize{67.1} & \scriptsize{13.4} & \scriptsize{81.8} & \scriptsize{8.19} & \scriptsize{90.7} & \scriptsize{9.47} & \scriptsize{89.0} & 2.94 \\

\textbf{Ours} & \textbf{2.92} & \textbf{98.6} & 3.94 & 98.0 & \textbf{2.69} & \textbf{99.2} & \textbf{2.74} & 97.9 & \textbf{0.94} & \textbf{100} & \textbf{13.0} & \textbf{83.2} & \textbf{8.40} & \textbf{92.1} & \underline{3.16} & \underline{97.5} & \textbf{4.72} & \textbf{95.8} & \textbf{1.56} \\
        
        \hline  
        \multicolumn{19}{c}{Affine-invariant disparity}\\
        \hline
        
LeReS & 7.31 & 95.5 & 12.2 & 87.1 & 10.2 & 90.1 & 8.44 & 92.9 & 4.33 & 99.7 & 28.9 & 59.6 & 23.4 & 73.0 & 10.7 & 88.3 & 13.2 & 85.8 & 8.25 \\

ZoeDepth & \textcolor{lightgray}{5.21} & \textcolor{lightgray}{97.7} & \textcolor{lightgray}{5.84} & \textcolor{lightgray}{95.6} & 8.07 & 94.0 & 6.19 & 96.1 & 2.60 & 99.9 & 26.9 & 66.3 & 14.1 & 81.7 & 8.17 & 92.0 & \textcolor{lightgray}{9.63} & \textcolor{lightgray}{90.4} & 6.75 \\

DUSt3R & 4.24 & 98.1 & 7.72 & 92.1 & 5.60 & 96.2 & 4.49 & 96.6 & 2.63 & 99.8 & 40.0 & 56.7 & 17.4 & 76.2 & 7.10 & 92.8 & 11.1 & 88.6 & 6.75 \\

Metric3D V2 & 13.4 & 81.5 & \underline{3.76} & \underline{98.2} & \underline{4.30} & 97.7 & 8.55 & 92.3 & 1.80 & \textbf{100} & 21.8 & 72.4 & \textcolor{lightgray}{7.35} & \textcolor{lightgray}{94.1} & 7.70 & 90.2 & \textcolor{lightgray}{8.58} & \textcolor{lightgray}{90.8} & 5.29 \\

UniDepth V1 & \underline{3.78} & \textbf{98.7} & \textbf{3.64} & \textbf{98.7} & 5.34 & 97.2 & 4.06 & \underline{98.1} & 2.56 & 99.9 & 28.6 & 60.7 & \underline{9.94} & \underline{89.1} & 5.95 & 95.5 & 7.98 & 92.2 & 3.62 \\

MiDaS V3.1 & 4.58 & 98.1 & 6.25 & 94.7 & 5.77 & 96.8 & 4.73 & 97.4 & 1.86 & \textbf{100} & 21.3 & 73.1 & 14.5 & 82.6 & 6.05 & 94.9 & 8.13 & 92.2 & 5.00 \\

DA V1 & 4.20 & 98.4 & 5.40 & 97.0 & 4.68 & \underline{98.2} & 4.18 & 97.6 & 1.54 & \textbf{100} & \underline{20.1} & \underline{77.6} & 12.7 & 86.9 & 5.69 & 95.7 & \underline{7.31} & \underline{93.9} & \underline{3.00} \\

DA V2 & 4.14 & 98.3 & 5.61 & 96.7 & 4.71 & 97.9 & \underline{3.47} & \textbf{98.5} & \underline{1.24} & \textbf{100} & 21.4 & 72.8 & 13.1 & 86.4 & \underline{5.29} & \underline{96.1} & 7.37 & 93.3 & 3.12 \\

\textbf{Ours} & \textbf{3.38} & \underline{98.6} & 4.05 & 98.1 & \textbf{3.11} & \textbf{98.9} & \textbf{3.23} & 98.0 & \textbf{0.96} & \textbf{100} & \textbf{18.4} & \textbf{79.5} & \textbf{8.99} & \textbf{91.5} & \textbf{3.98} & \textbf{97.2} & \textbf{5.76} & \textbf{95.2} & \textbf{1.44} \\

        \specialrule{.12em}{0.12em}{0em}
    \end{tabular}
    \caption{Full table of comparison for depth map estimation. * methods have multiple model versions available for respective benchmarks, among which the best for each benchmark is chosen for ranking, followed by the detailed results in smaller text size for each version. \textcolor{lightgray}{Gray numbers} denote models trained on respective benchmarks.}
    \label{tab:comparison_depth_full}
\end{table*}

\paravspace
\paragraph{More qualitative comparisons}

\begin{figure*}[ht!]
    \centering
    \includegraphics[width=\linewidth]{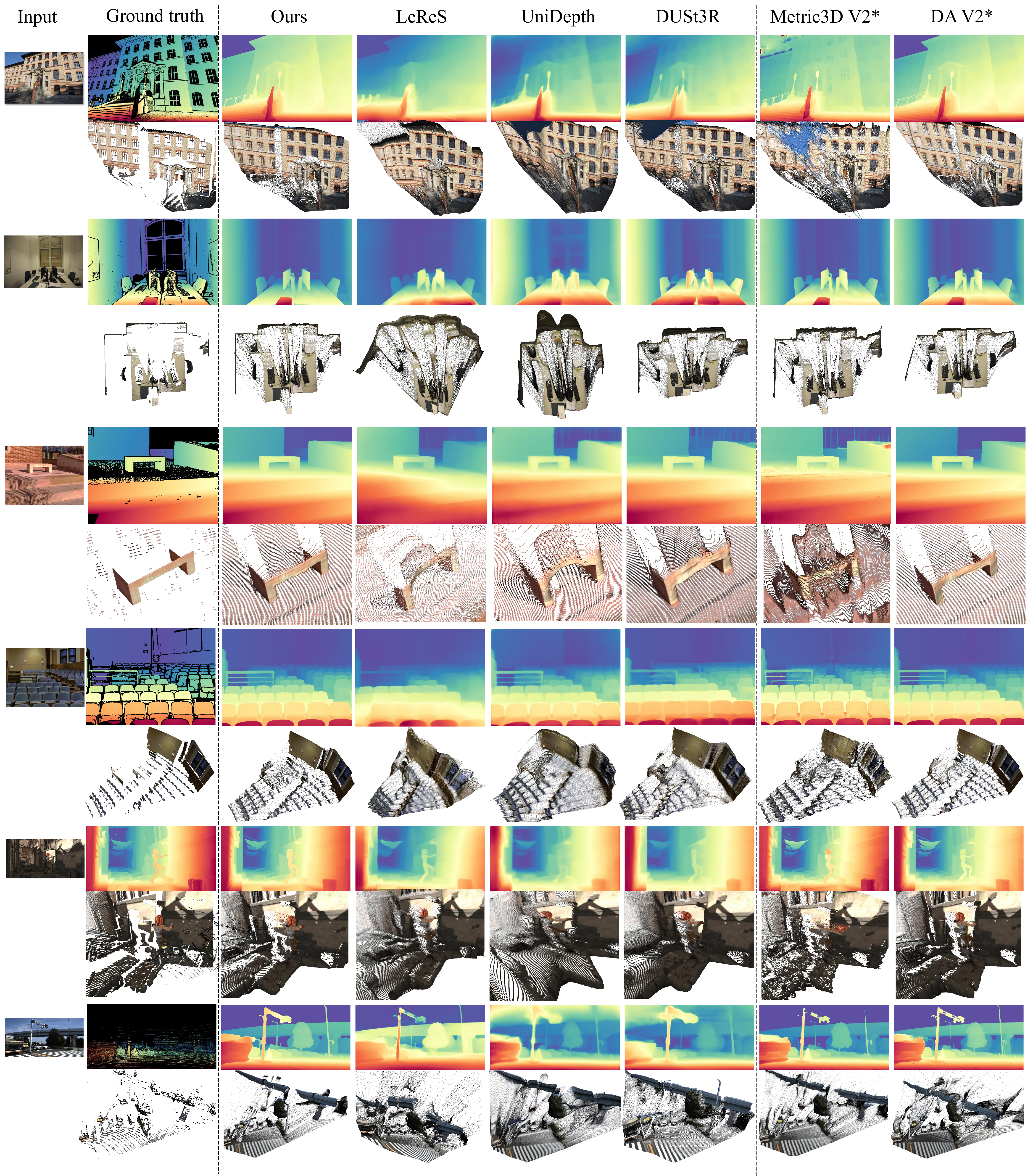}
    \caption{Additional qualitative comparisons from the \emph{evaluation datasets}. *: for methods without camera intrinsics prediction, ground-truth camera intrinsics (and disparity shifts) were used to lift their results into 3D points. \textbf{\emph{Best viewed with zoom.}}}
    \label{fig:more_qualitative_comparison_test}
\end{figure*}

\begin{figure*}[ht!]
    \centering
    \includegraphics[width=\linewidth]{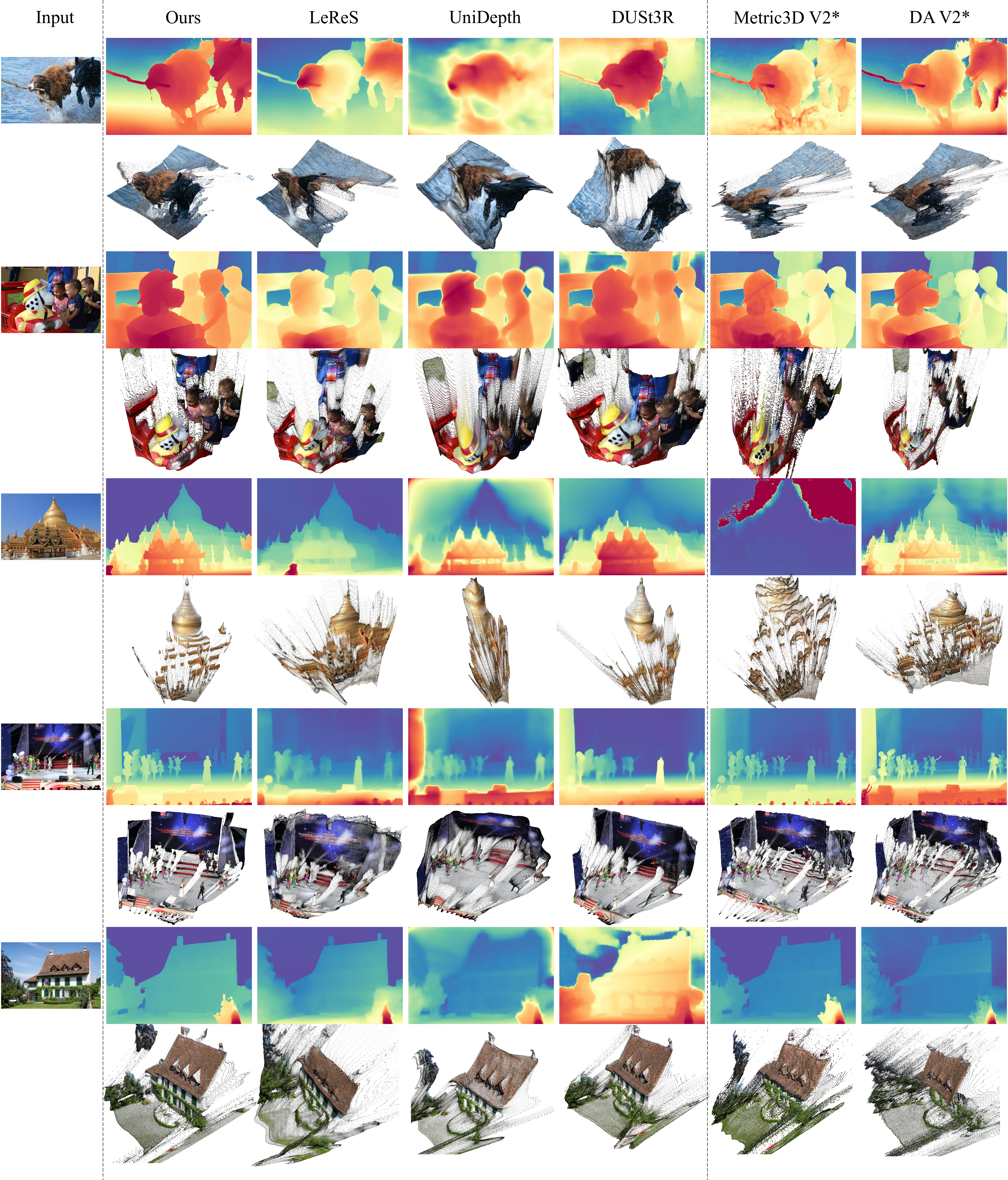}
    \caption{Additional qualitative comparisons for \emph{in-the-wild} images from the DIV2K dataset. *: for methods without camera intrinsics prediction, our camera intrinsics prediction were used to lift their results into 3D points. Supplementary videos contain more results. \textbf{\emph{Best viewed with zoom.}}}
    \label{fig:more_qualitative_comparison_wild}
\end{figure*}

Fig.~\ref{fig:more_qualitative_comparison_test} and Fig.~\ref{fig:more_qualitative_comparison_wild} present additional visual comparisons on zero-shot evaluation datasets and in-the-wild images. Our method is compared with LeReS~\cite{wei2021leres}, UniDepth~\cite{piccinelli2024unidepth}, DUSt3R~\cite{wang2024dust3r}, Metric3D V2~\cite{hu2024metric3dv2}
and Depth Anything V2~\cite{yang2024depthanythingv2}. Since Metric3D V2 and Depth Anything V2 predict depth map and require ground truth camera focal to obtain 3D points cloud results, we visualize them using our estimated focal lengths. 

In the supplementary videos, we present \emph{extensive and uncurated comparisons} using the first 100 images from the DIV2K\cite{div2kworkshop} dataset.

\paravspace
\paragraph{More visual results}

\begin{figure*}[ht!]
    \centering
    \includegraphics[width=0.88\linewidth]{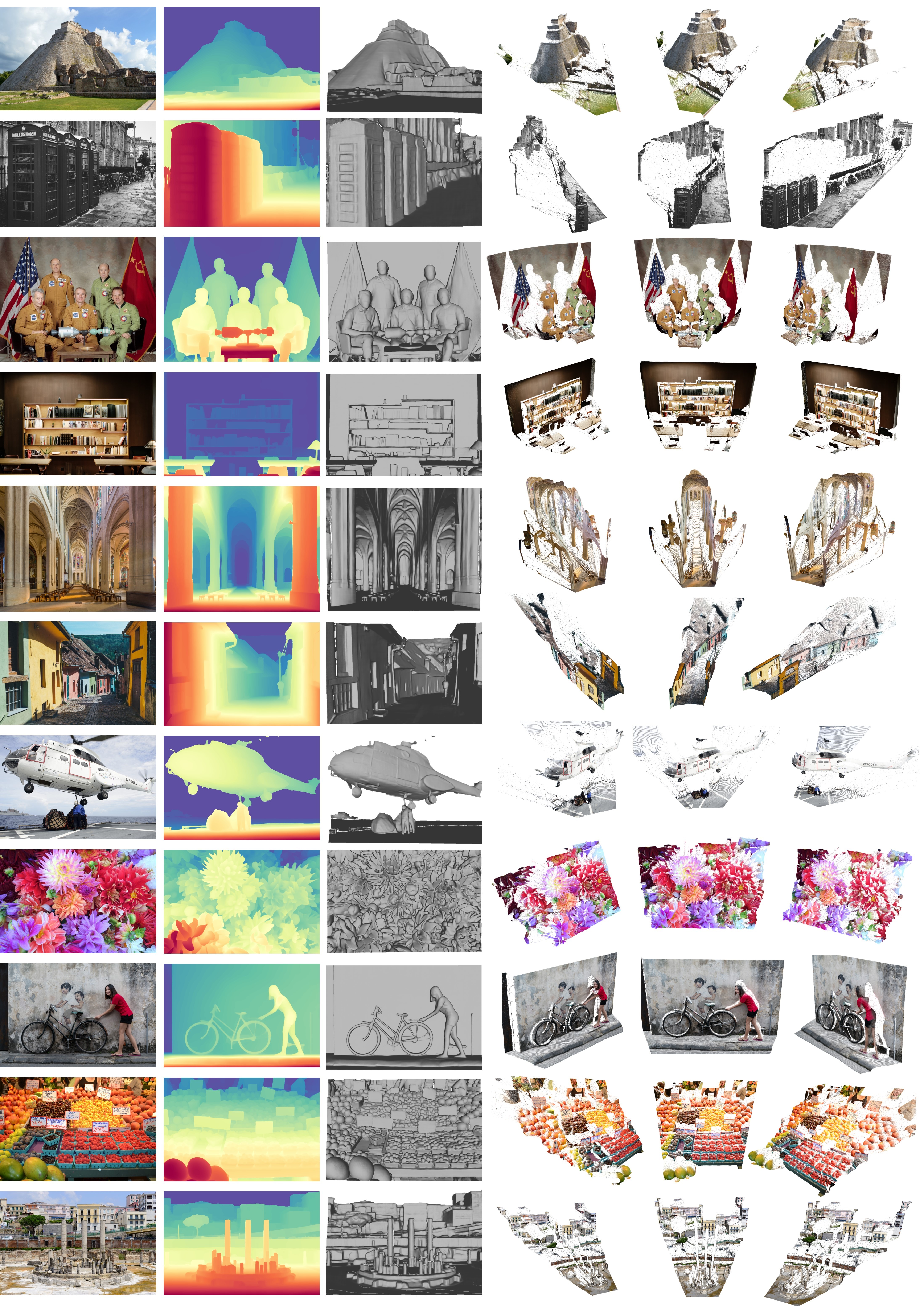}
    \vspace{-1em}
    \caption{Additional visual results for open-domain images of our model (page 1 of 2). The columns from left to right are the input images, reconstructed disparity maps, reconstructed surface geometry viewed from the source view, and three novel-view images, respectively.}
    \label{fig:more_results_1}
\end{figure*}

\begin{figure*}[ht!]
    \centering
    \includegraphics[width=0.88\linewidth]{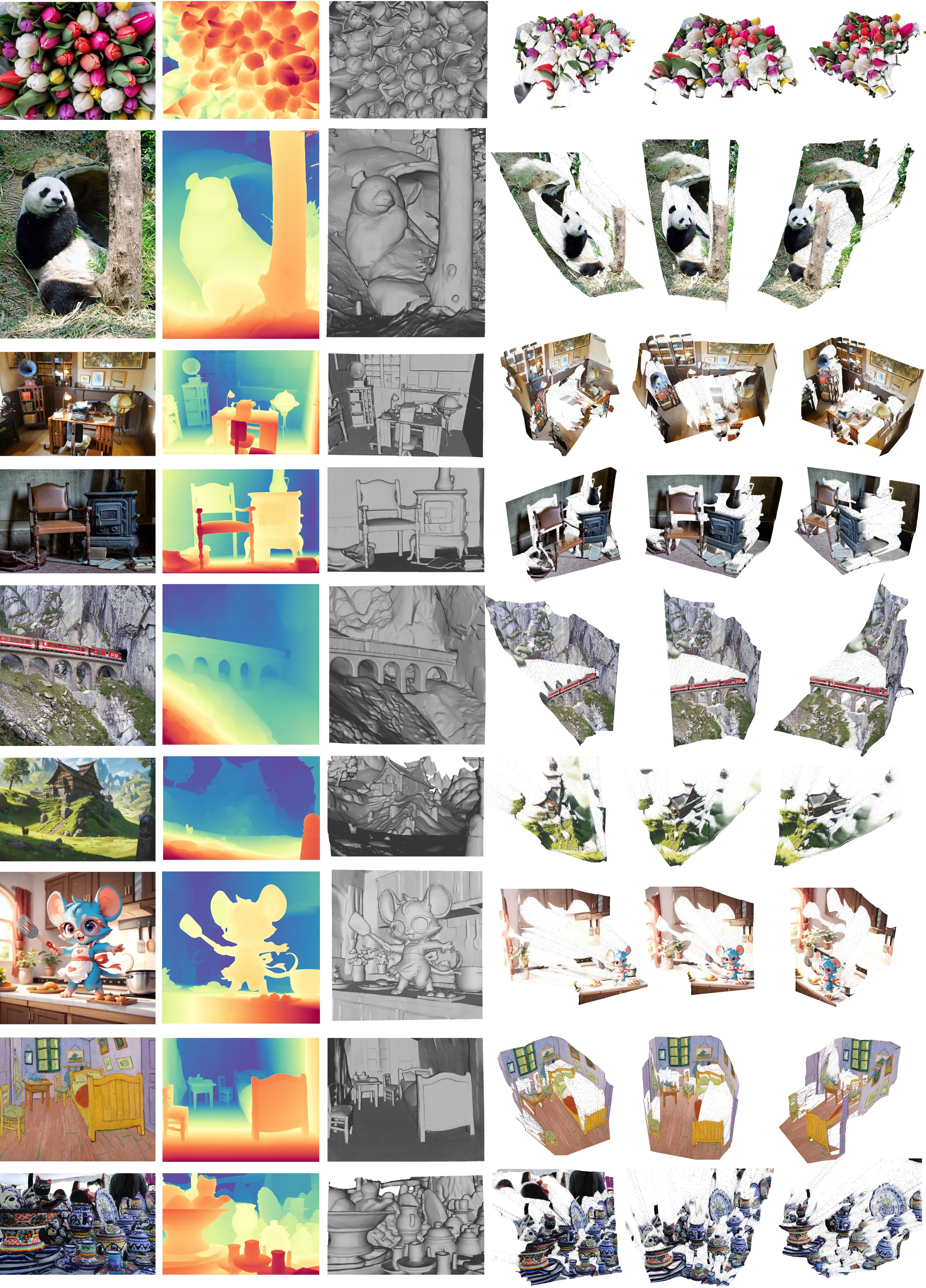}
    \vspace{-1em}
    \caption{Additional visual results for open-domain images of our model (page 2 of 2). The columns from left to right are the input images, reconstructed disparity maps, reconstructed surface geometry viewed from the source view, and three novel-view images, respectively.}
    \label{fig:more_results_2}
\end{figure*}

In Fig.~\ref{fig:more_results_1} and Fig.~\ref{fig:more_results_2}, we demonstrate more reconstruction results of our method for more open-domain images. 

\section{Limitations and Future Work}
\begin{figure}[H]
    \centering
    \includegraphics[width=\linewidth]{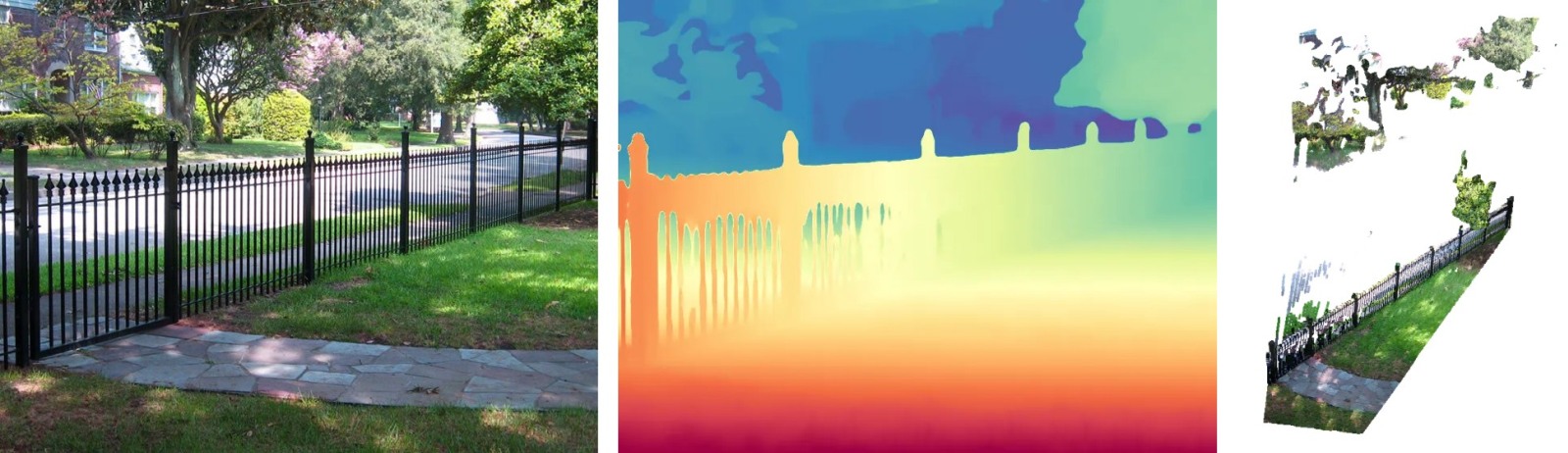}
    \caption{A failure case. Our model fails to capture the thin structure of the fence, leading to a flattened geometry.}
    \label{fig:failure_case}
\end{figure}

While our model demonstrates strong performance, accurately capturing thin structures remains a significant challenge. This difficulty arises from the network's limited capacity and the presence of noisy real-world training data. As illustrated in Fig.~\ref{fig:failure_case}, our model may fail to recover these intricate structures.

Additionally, while monocular video reconstruction holds great promise as an application, achieving temporal consistency presents substantial challenges. Our model, designed for single-image input, cannot inherently maintain temporal coherence due to the ambiguity of the task. Addressing this issue would require non-trivial solutions, such as global optimization techniques.
Given the rapid advancements in video depth estimation, we believe that an end-to-end model for monocular video reconstruction could significantly benefit from our proposed techniques. Exploring this direction is a compelling avenue for future work.

\end{document}